\documentclass[11pt,a4paper]{article}
\usepackage{times,latexsym}
\usepackage{url}
\usepackage[T1]{fontenc}

\usepackage[inline]{enumitem}
\usepackage{hyperref}
\usepackage{booktabs}
\usepackage{graphicx}
\usepackage{multirow}
\usepackage{multicol}
\usepackage{algorithm}
\usepackage{algpseudocode}
\usepackage{etoolbox}
\AtBeginEnvironment{algorithmic}{\scriptsize}

\usepackage{float}
\usepackage{listings}
\usepackage{caption}
\usepackage{amssymb}
\usepackage{amsmath}
\usepackage{cleveref}
\usepackage{pifont}
\usepackage{xcolor}

\usepackage[acceptedWithA]{tacl2021v1}

\usepackage{xspace,mfirstuc,tabulary}

\newif\iftaclinstructions
\taclinstructionsfalse 
\iftaclinstructions
\renewcommand{\confidential}{}
\renewcommand{\anonsubtext}{(No author info supplied here, for consistency with
TACL-submission anonymization requirements)}
\newcommand{\instr}
\fi

\iftaclpubformat 

\else

\fi

\newcommand{\TeQoDO}{TeQoDO}

\newcommand{\simfunction}{\operatorname{sim}}
\newcommand{\DB}{\textup{DB}}

\newcommand{\predsubscript}{\textup{pred}}
\newcommand{\simsubscript}{\textup{sim}}
\newcommand{\truesubscript}{\textup{true}}

\title{Text-to-SQL Task-oriented Dialogue Ontology Construction}

\author{Renato Vukovic, Carel van Niekerk, Michael Heck, Benjamin Ruppik, \\{\bf  Hsien-chin Lin, Shutong Feng, Nurul Lubis, Milica Ga\v{s}i\'{c}} \\
  Heinrich Heine University Düsseldorf, Germany\\
  \texttt{\small{\{revuk100,niekerk,heckmi,ruppik,linh,fengs,lubis,gasic\}@hhu.de}} \\}

\date{}

\begin{document}

\maketitle
\begin{abstract}
    Large language models (LLMs) are widely used as general-purpose knowledge sources, but they rely on parametric knowledge, limiting explainability and trustworthiness.
    In task-oriented dialogue (TOD) systems, this separation is explicit, using an external database structured by an explicit ontology to ensure explainability and controllability.
    However, building such ontologies requires manual labels or supervised training. \\
    We introduce \textbf{TeQoDO}: a \textbf{Te}xt-to-S\textbf{Q}L task-\textbf{o}riented \textbf{D}ialogue \textbf{O}ntology construction method. 
    Here, an LLM autonomously builds a TOD ontology from scratch using only its inherent SQL programming capabilities combined with concepts from modular TOD systems provided in the prompt. \\
    We show that TeQoDO outperforms transfer learning approaches, and its constructed ontology is competitive on a downstream dialogue state tracking task.
    Ablation studies demonstrate the key role of modular TOD system concepts.
    {\TeQoDO} also scales to allow construction of much larger ontologies, which we investigate on a Wikipedia and arXiv dataset.
    We view this as a step towards broader application of ontologies. 
    \footnote{Code is available under \url{https://gitlab.cs.uni-duesseldorf.de/general/dsml/teqodo-code-public}}
\end{abstract}

\section{Introduction}
\label{sec:intro}

Large language models (LLMs) have become ubiquitous knowledge processing systems, with the ability to reach or surpass human performance on a wide variety of natural language processing tasks.
These models are pre-trained on massive corpora and aligned via human feedback using reinforcement learning \cite{brown2020gpt3, ouyang2022instructgpt}.
Despite these remarkable abilities, there are some inherent problems associated with these systems. 
Their knowledge is stored in vast numbers of parameters, which makes it very difficult to understand their behaviour, even via probing \cite{cifka-liutkus-2023-black}.
It is therefore extremely challenging to verify their inherent knowledge \cite{zhong2024knowledge}.
Moreover, LLMs often produce plausible-looking hallucinations \cite{sahoo-etal-2024-comprehensive,feng-etal-2024-infusing}. 
Finally, many LLMs and their training data are closed-source.

\emph{Ontologies} provide a human-readable means of reconstructing knowledge-based reasoning in language models (LMs) by capturing knowledge in a structured
and explicit manner \cite{gruber1995ontologydesign,lo2024ontologylearning}.
Manually building ontologies for all relevant domains is infeasible \cite{milward2003ontology}.
To ensure broad applicability, ontologies should be constructed automatically and consistently.
Automatic ontology construction is a promising solution to these challenges.

\begin{figure}[t]
    \centering
    \includegraphics[width=0.8\linewidth]{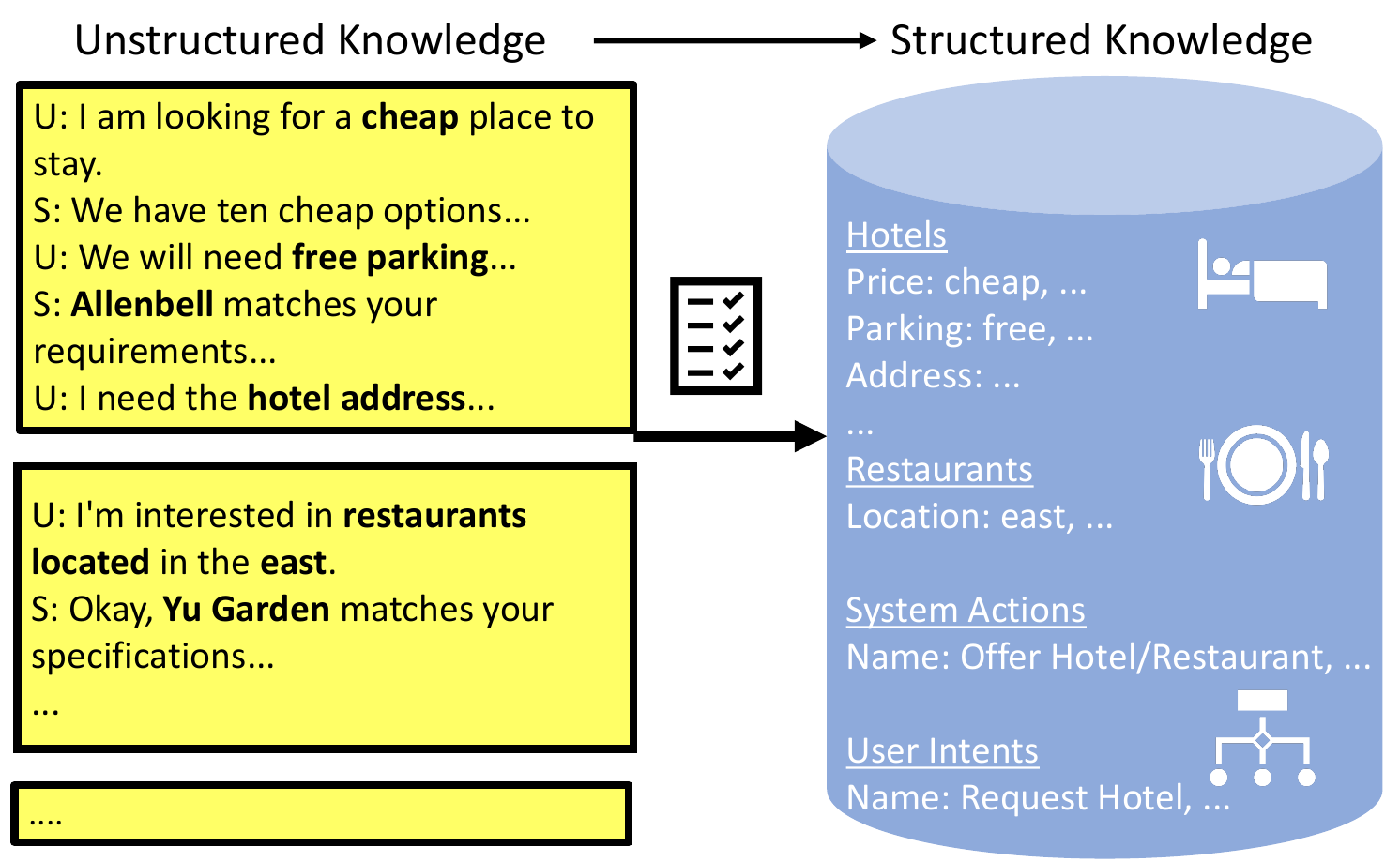}
    \caption{
        Example ontology construction shows the extraction of the domain ``Hotels'', with slot ``Price'' and value ``cheap''.
        Actions and intents are defined based on the domains and slots.
        \label{fig:ontology_construction_intro}
    }
    \vspace{-15pt}
\end{figure}

In this work, we focus on constructing task-oriented dialogue (TOD) ontologies from dialogues.
The goal is to extract information and organise it into a hierarchy for handling user queries.
TOD ontologies consist of \emph{domains}, \emph{slots}, and \emph{values}, with \emph{system actions} and \emph{user intents}.
\Cref{fig:ontology_construction_intro} illustrates ontology construction in this setting.
Ontologies are essential for TOD systems \cite{young2013pomdp,hudecek-dusek-2023-large}.

Most TOD ontology construction approaches follow two steps \cite{hudecek-etal-2021-discovering,vukovic-etal-2022-dialogue}: 
(1) term extraction from dialogue data and (2) relation extraction between the terms from the previous step.
Existing approaches tackle these steps separately and rely on annotated training data \cite{finch-etal-2024-transforming,vukovic-etal-2024-dialogue}.
This has the downside of additional and potentially error-prone training, and the danger of information loss between the two processing steps.

We propose \textbf{{\TeQoDO}}, a \textbf{Te}xt-to-S\textbf{Q}L task-\textbf{o}riented \textbf{D}ialogue \textbf{O}ntology construction approach.
\Cref{fig:TeQoDOmain} provides an overview of the method.
{\TeQoDO} uses LLMs’ code understanding and generation capabilities \cite{chen2022program} to build ontologies from scratch via SQL.
It incrementally constructs the database by iterating over dialogues, retrieving relevant existing content, and updating the DB with new information.
The model uses state tracking from modular TOD to distinguish new content from existing data in the DB.
The update prompt includes a notion of success to help align schema changes with the user’s goals.
Query results are augmented with semantically similar concepts or example values from existing columns to improve coverage and consistency.

Text-to-SQL provides a structured format familiar to LLMs through code encountered during pre-training \cite{deng-etal-2022-recent,zhao-etal-2024-deciphering}.
It reduces the need for textual prompts, as SQL tables, columns, and values align with TOD ontology domains, slots, and values.
To our knowledge, we are the first to use text-to-SQL to build a TOD ontology from scratch.
We also adapt similarity-based evaluation metrics for ontology learning \cite{lo2024ontologylearning} to reflect the hierarchical nature of TOD ontologies.
This enables evaluation of various concept types, such as domains and slots.

We run experiments on two widely used TOD datasets: MultiWOZ \cite{eric-etal-2020-multiwoz} and Schema-Guided Dialogue (SGD) \cite{rastogi2020towards}.
{\TeQoDO} outperforms recent TOD ontology construction methods \cite{finch-etal-2024-transforming,vukovic-etal-2024-dialogue}.
To test generalisation, we apply {\TeQoDO} to general ontology datasets from \citet{lo2024ontologylearning} and show it scales to large ontologies.
In summary, our contributions are:

\begin{itemize}
    \item We propose a text-to-SQL framework for building TOD ontologies from scratch, allowing LLMs to update a DB with each newly observed dialogue without supervision or rule-based aggregation.
    \item The constructed ontology enables dialogue state tracking with performance comparable to using the ground truth ontology.
    \item {\TeQoDO} generalises to large-scale ontology datasets like Wikipedia and arXiv, performing competitively.
\end{itemize}

\section{Related Work}
\label{sec:related_work}

\subsection{Ontology Construction}
\label{subsec:ontology_construction}

Traditionally, ontologies are handcrafted.
Prior to deep learning, efforts for automatic ontology construction relied on frequency-based rules and linguistic features.
With LMs, clustering and supervised methods enabled feature-based learning.

\paragraph{Rule-based} ontology construction \citep{frantzi1999c,nakagawa2002simple} relies on frequency-based methods to extract key subsequences from text.
\citet{wermter2006you} use linguistic features for more advanced term extraction.
While interpretable, these methods are hard to adapt across domains and often yield low precision.

\paragraph{Clustering-based} approaches, such as \citet{yu-etal-2022-unsupervised}, apply unsupervised parsing and hierarchical clustering to group extracted terms into domains and slots.
\citet{finch-etal-2024-transforming} train a generative slot induction model to extract domain-slot-value triples, which are then clustered into a slot hierarchy.
These methods usually produce many slots, whose interpretation depends entirely on how well they match the ground truth. 
They also rely heavily on data quality, embedding representations, and sensitive hyperparameters, such as the minimum cluster size.

\paragraph{Supervised}
\citet{lo2024ontologylearning} fine-tune LLMs to predict ontology sub-graphs at the document level and merge them using frequency-based rules to construct ontologies for Wikipedia and arXiv datasets.
They also propose evaluation functions covering both verbatim content and higher-level structural semantics compared to the ground truth, which we adapt to TOD.
\citet{vukovic-etal-2024-dialogue} train a model to predict TOD ontology subgraphs at the dialogue level and aggregate them by merging all predictions.
They improve generalisability by updating the decoding with constraints and confidence-based adjustments.
However, their method focuses only on relation extraction, using ground truth terms as input.
These methods require costly annotated training data from \cite{budzianowski-etal-2018-multiwoz}.
In contrast, our approach requires no training and conducts both term and relation extraction.
We use \citet{vukovic-etal-2024-dialogue} and \citet{finch-etal-2024-transforming} as baselines.

\subsection{Text-to-SQL}

\citet{li2023llmDBinterface} present a large benchmark for text-to-SQL queries on big databases and find LLMs underperform compared to humans.  
\citet{zhou_db-gpt_2024} apply LLMs to database tasks such as query rewriting and index tuning using automatic prompt generation, DB-specific pre-training, model design, and fine-tuning.  
\citet{yang-etal-2024-synthesizing} enhance text-to-SQL parsing for smaller models by synthesising training data with larger models.  
\citet{allemang2024ontologies} improve text-to-SPARQL \cite{Polleres2014sparql} using a rule-based query checker that leverages the ontology, with an LLM repairing queries based on this feedback.  
\citet{stricker-paroubek-2024-shot} use LLMs in a few-shot TOD setup by translating user queries into SQL to interact with tables in the underlying ontology.  
In contrast, we build a DB and ontology from scratch using text-to-SQL.

\subsection{Dialogue Flow Induction}

\citet{burdisso-etal-2024-dialog2flow} propose Dialog2Flow (D2F) embeddings that map utterances to a latent space according to the actions they represent.
These actions are relevant for dialogue flow and help unify action representations across datasets.
\citet{agrawal-etal-2024-dialog} induce dialogue flows unsupervisedly via clustering and merging, ensuring relevance, coverage, clarity, and non-redundancy.
\citet{choubey-etal-2025-turning} retrieve relevant conversations and generate dialogue workflows using a question–answer chain-of-thought prompt with an LLM, then evaluate them via dialogue simulation with workflow decomposition.
These methods focus on inducing general system and user actions, closely related to intent and action prediction in {\TeQoDO}.
However, our approach also induces a fine-grained domain–slot–value hierarchy linking actions to underlying structured knowledge.



\section{Text-to-SQL Ontology Construction}
\label{sec:method}

\begin{figure}[t]
    \centering
    \includegraphics[width=\linewidth]{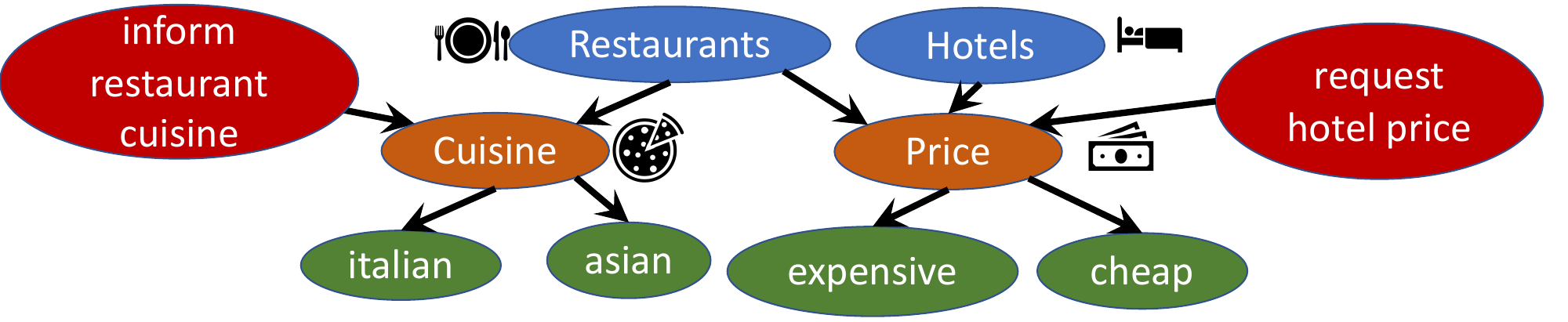} %
    \caption{
        Example TOD ontology.
        \label{fig:ontology_example}
    }
    \vspace{-15pt}
\end{figure}

\begin{figure*}[t]
    \centering
    \includegraphics[width=0.7\linewidth]{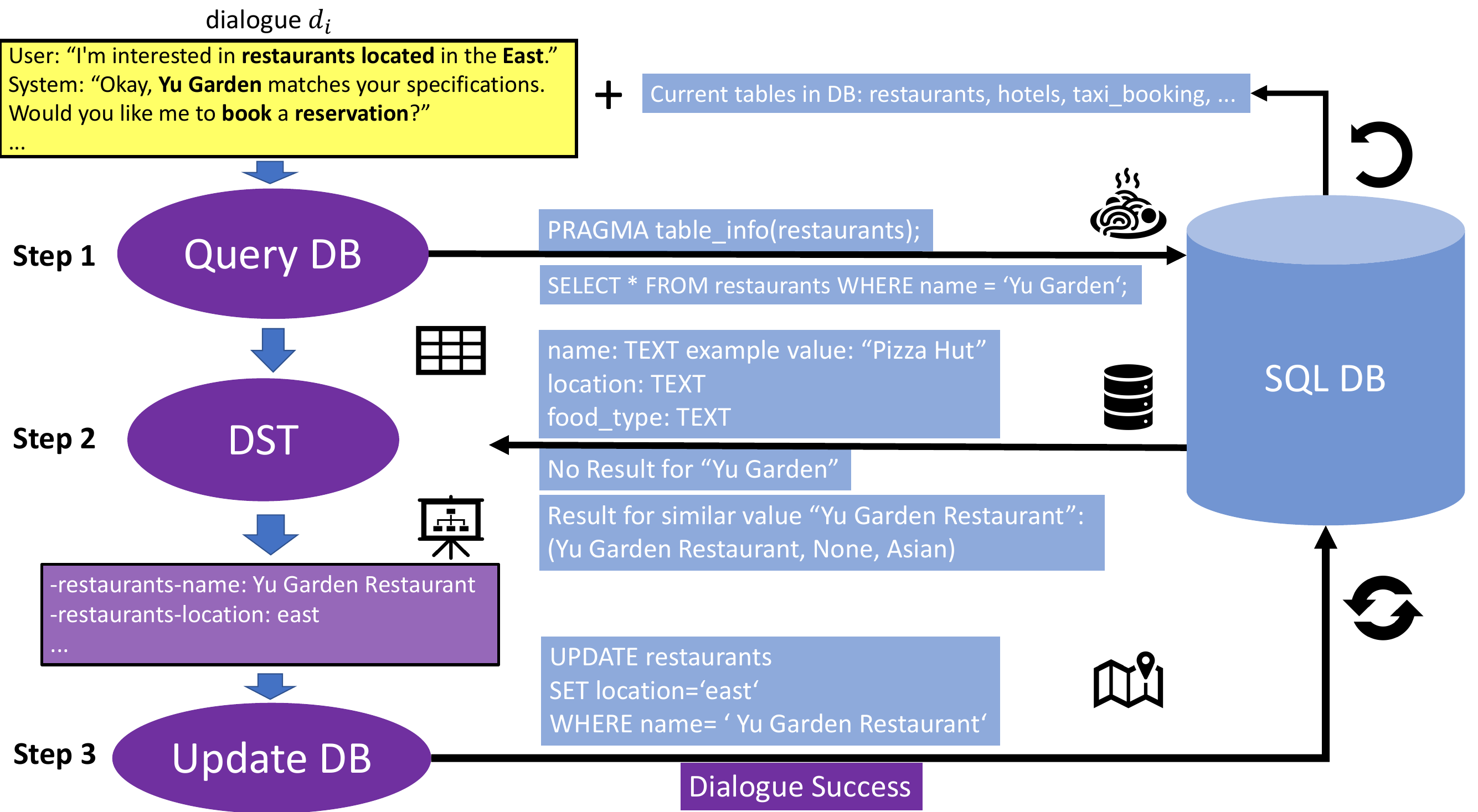} %
    \caption{
        {\TeQoDO} Overview with example DB queries and results.
        \label{fig:TeQoDOmain}
    }
    \vspace{-15pt}
\end{figure*}

\subsection{Problem Formulation} 
\label{subsec:TeQoDO:problem}

Given a \emph{TOD dataset} $D = \{d_1, \ldots, d_k\}$, each dialogue \(d_{i}\) is comprised of alternating user and system turns \(\{u_1, s_1, \ldots, u_{m_{i}}, s_{m_{i}}\}\) containing information about user-queried entities.
The goal is to induce an ontology $O_D$ for the dataset.
The \emph{ontology} is a directed graph $O_D = (V, E)$ with five node types $V = V_\text{domain} \cup V_\text{user\_intent} \cup V_\text{system\_action} \cup V_\text{slot} \cup V_\text{value}$, forming a three-level hierarchy.
The edge set \(E\) is a subset of edges between specific node types,
\(
E \subseteq (V_\mathrm{domain} \times V_\mathrm{slot}) \cup (V_\mathrm{slot} \times V_\mathrm{value}) \cup (V_\mathrm{user\_intent} \times (V_\mathrm{domain} \times V_\mathrm{slot})) \cup (V_\mathrm{system\_action} \times (V_\mathrm{domain} \times V_\mathrm{slot})).
\)

See \Cref{fig:ontology_example} for an example TOD ontology, with domains in blue, slots in brown, values in green, and intents and actions in red.
Domains represent broad topics, while slots specify particular types of information.
Values provide concrete content for the slots.
User intents and system actions define how domain-slot pairs can be used, based on the domain-slot-value hierarchy.



\subsection{Method} 
\label{subsec:TeQoDO:method}

Our main prompting approach combines TOD modelling with SQL, as outlined below.
We then detail the steps of our iterative ontology construction pipeline.
We utilise an LLM that creates SQL queries, which are then executed.

\paragraph{SQL-Background}

SQL is a query language for relational databases \cite{chamberlin1974sql}.
We use SQLite,\footnote{\label{footnote:sqlite}\url{https://www.sqlite.org}} a serverless relational DB management system suitable for small-scale databases and schema-based knowledge representation.
SQL stores data in tables with columns and values, e.g., a ``restaurant'' table with a ``price'' column and ``expensive'' as a value.
This structure aligns with TOD ontology formats, making it well-suited for ontology construction, as it reduces the prompt engineering needed for the LLM to induce the TOD ontology structure.
For intents and actions, there are separate tables that store names of intent and action entries.

{\TeQoDO} primarily uses data retrieval, definition, and manipulation queries.
To access current DB content, a \texttt{SELECT} query retrieves table names.
\texttt{PRAGMA} queries fetch column names and data types for each table.
\texttt{SELECT} can also query specific values from tables and columns.
Tables are created with \texttt{CREATE TABLE}, specifying column names and types.
\texttt{ALTER TABLE} adds columns, \texttt{INSERT INTO} adds entities, and \texttt{UPDATE} modifies column values.

\paragraph{Task-Oriented Dialogue Modelling}

We incorporate two concepts from modular task-oriented dialogue models \citep{young2013pomdp} into our prompt to improve the quality of generated update queries.
First, \emph{dialogue state tracking} (DST) is used as a separate step to distinguish existing database content from new input, based on the current schema.
Second, we prompt the model to consider \emph{dialogue success}, guiding it to make updates that support achieving the user goal.
For a more formal description, see \Cref{appendix:tab:TeQoDO_mathtable} in the appendix.

\paragraph{{\TeQoDO} Prompting Steps}
See \Cref{fig:TeQoDOmain} for an overview of {\TeQoDO} with example queries and results.
The pipeline steps are given in \Cref{alg:TeQoDO} and are described in the following.

\paragraph{Step 1: Query existing DB information}

To ensure structural and naming consistency in {\TeQoDO}, the model is first prompted to query the current DB tables that are relevant for the current input dialogue \(d_i\).
This is done by giving the model the whole set of current tables in the DB as an additional input.
Initially, as the ontology is built from scratch, the DB is empty, and thus no results are returned.
The resulting table list $T = \{t_1, \ldots, t_n\}$ is included with the dialogue.
The model is prompted to query column info for tables relevant to the dialogue.
We then append the column query results to the prompt so the model generates \texttt{SELECT} queries consistent with the DB state.
Along with the results of the column information queries subsamples of explicit values a column contains are given to the model.
In the final part of this step, it is prompted to generate \texttt{SELECT} queries to retrieve specific entity values from the dialogue.
To reduce variation in column values (e.g. ``Alexander Bed and Breakfast'' vs ``Alexander B \& B'') and improve naming consistency in the DB, we also experiment with query results for similar tables, columns, and values.
This decreases duplicate entries by increasing the chance of reusing existing values.
In the final {\TeQoDO} pipeline, we only utilise the column value subsamples.

\paragraph{Step 2: Dialogue State Tracking with DB information}

In this step, the model restructures DB query results into DST style labels.
It predicts the current dialogue’s value for each table-column pair using the DB’s stored information.
This improves the distinction between existing DB information and new information from the dialogue.

The model is prompted to summarise dialogue states using the DB’s table and column names.
This summary includes information from both the user and the system.
By predicting the state, the model condenses the dialogue’s information into a concise, model-determined format not fully detailed in the prompt.
Notably, the model tracks the entire dialogue at once, not turn-by-turn.
It must track all the mentioned information, not only what the user queries.
Since the input is a whole dialogue, the model does not do traditional DST, but rather tracks the information that was mentioned in the dialogue in a format that is based on the current DB schema.
The format here resembles DST labels with slot-value information.

\paragraph{Step 3: DB Update with Success}

In the final step, the model generates database update queries based on the current DB state and dialogue information.
It is prompted to create update queries considering DB query and DST results.
Update queries include \texttt{CREATE TABLE}, \texttt{ALTER TABLE} to add columns, and \texttt{UPDATE} to add values.
The model must follow the existing DB structure and generate consistent updates with existing tables, not create new ones.
Additionally, the model is prompted to update the DB to support fulfilling the user’s goal by incorporating dialogue success into the prompt.
Namely, this instruction is added to the update prompt: ``\textit{so that the user’s goal expressed in the dialogue can be successfully fulfilled using only information stored in the database.}''

\section{Experiments}
\label{sec:experiments}

\subsection{Datasets}
\label{subsec:exp:datasets}

\paragraph{MultiWOZ}
The first dataset we employ is MultiWOZ 2.1 \citep{eric-etal-2020-multiwoz}, a large-scale, multi-domain dialogue dataset containing human-human conversations annotated with domain, slot, and value labels. 
It covers information such as hotel bookings, restaurant reservations, taxi services, and attractions.
We utilise the test set with \(1,000\) dialogues and \(6\) domains.

\begin{algorithm}[t]
    \scriptsize
    \caption{
        {\TeQoDO}
    }
    \label{alg:TeQoDO}
    \begin{algorithmic}[1]
    \State \textbf{Input:} Dialogue dataset $D$, Existing Database $\DB$, Prompt $p_0$
    \For {$d_i$ in $D$}
        \State Query current set of tables $T = \{t_1, ..., t_n\} \in \DB$:
        \State Query table columns $c_{T,d_i}$ using prompt $p = p_0 + d_i + T$
        \State Generate \texttt{SELECT} queries $v_{T,d_i}$ using $p = p + c_{T,d_i}$
            \State Track information $ST_{d_i}$ using $p = p + v_{T,d_i}$
        \State Generate update queries $U_{d_i}$ for success  using $p = p + ST_{d_i}$
    \EndFor
    \end{algorithmic}
\end{algorithm}

\paragraph{SGD}
Second, we use the schema-guided dialogue (SGD) dataset \citep{rastogi2020towards}.
It is a diverse, multi-domain dataset for TOD modelling, featuring detailed schema annotations.
It covers services like flight booking, calendar scheduling, banking, and media services, including unseen services at test time.
We use the SGD test split for evaluation in the main results, containing $4,201$ dialogues and $18$ domains.
We load these with ConvLab-3~\cite{zhu-etal-2023-convlab-local}.

\paragraph{General Ontology Data}
To test whether {\TeQoDO} applies beyond TOD data, we use the Wikipedia and arXiv datasets by \citet{lo2024ontologylearning}.
The Wikipedia test set contains $242,148$ titles and abstracts; its gold ontology has four hierarchy levels with $8,033$ nodes and $14,673$ edges.
Topics include, for example, types of \textit{``injuries''} or \textit{``legislative bodies''}.
Here, the input to {\TeQoDO} are titles and summaries of Wikipedia articles rather than task-oriented dialogues.
\citet{lo2024ontologylearning} construct the dataset by running a breadth-first search from the root category ``Main topic classifications'' to depth 3, collecting titles and lead summaries of up to $5,000$ pages per category using the Wikipedia API\footnote{\url{https://en.wikipedia.org/w/api.php}}.

The arXiv test set has $27,630$ title–abstract pairs, two hierarchy levels with $61$ nodes and $61$ edges.
Main topics are arXiv categories such as \textit{``Mathematics''} with subcategories like \textit{``Commutative Algebra''}, making this ontology more abstract and higher level than Wikipedia’s.
Here, the input to {\TeQoDO} are titles and abstracts of arXiv papers rather than TODs.
The taxonomy is taken from the official site, and the corpus includes titles and abstracts of all arXiv papers from 2020–2022 with at least $10$ citations.
These ontologies do not follow the domain–slot–value hierarchy, making SQL less intuitive.

\subsection{Evaluation} 
\label{subsec:TeQoDO:eval}

We run {\TeQoDO} on a dataset, generating the ontology by executing all update queries per dialogue.
For evaluation, tables map to domains, columns to slots, and values to ground truth values.
System actions and user intents are matched by aligning table names with the ``system actions'' and ``user intents'' keys that are explicitly stored in the ground truth.
This means that the model is expected to predict one table for system actions and one for user intents.
The  values are then directly compared to the values stored under the keys in the ground truth ontology.

We adapt the general ontology evaluation framework of \citet{lo2024ontologylearning} to our task-oriented dialogue ontology structure, which includes domains, slots, values, system actions, and user intents. 
Specifically, we use \emph{literal} F1 as a hard metric and \emph{fuzzy} and \emph{continuous} F1 from \citet{lo2024ontologylearning} as soft metrics, as these capture most relevant evaluation aspects according to their results. 
The hard metric captures syntactic similarity, while the soft metrics capture some human intuition on semantic similarity.
Since these metrics treat all ontology graph edges equally -- overlooking infrequent higher-level relations -- we extend them to account for hierarchical levels and their specific edge types by evaluating the different levels separately.
We use all metrics for the ablation study and choose the continuous metric as our main metric, as it covers semantic similarity while being strict on the final structure of the ontology.
Note that we adopt the naming of the chosen metrics from \citet{lo2024ontologylearning} to remain aligned with established terminology from general ontology construction to facilitate a comparison.

We compute the macro average of each metric across the node classes: domains, slots, values, intents, and actions. 
Let $V_{i_{\text{\predsubscript}}}$ denote the predicted nodes and $V_{i_{\text{\truesubscript}}}$ the ground truth for class $i$.

\paragraph{Literal Metric}
Here, only exact matches count: true positives are $V_{i_{\predsubscript}} \cap V_{i_{\truesubscript}}$, false positives are $V_{i_{\predsubscript}} \setminus V_{i_{\truesubscript}}$, and false negatives are $V_{i_{\truesubscript}} \setminus V_{i_{\predsubscript}}$.

\paragraph{Fuzzy Metric}
In both the fuzzy and continuous setups, we map nodes above a cosine similarity threshold to ground truth nodes.
We use a sentence transformer \cite{reimers-gurevych-2019-sentence} and set the threshold at $t_{\textup{\simsubscript}} = 0.436$.\footnote{\label{footnote:threshold}This threshold corresponds to the median similarity for synonyms in WordNet \cite{miller-1994-wordnet,lo2024ontologylearning} using the ``all-MiniLM-L6-v2'' sentence-transformers model.}
For the fuzzy setup, all predicted nodes above the threshold are mapped to ground truth.
True positives are predicted nodes that are mapped to at least one ground truth node, i.e., \(\{v_p \in V_{i_{\predsubscript}} \mid \exists v_t \in V_{i_{\truesubscript}}: \simfunction(v_p, v_t) > t_{\simsubscript}\}\).
False positives are predicted nodes with similarity below or equal to the threshold for all ground truth nodes, i.e., \(\{v_p \in V_{i_{\predsubscript}} \mid \forall v_t \in V_{i_{\truesubscript}}: \simfunction(v_p, v_t) \leq t_{\simsubscript}\}\).
False negatives are ground truth nodes without any predicted node mapped above the threshold, i.e., \(\{v_t \in V_{i_{\truesubscript}} \mid \forall v_p \in V_{i_{\predsubscript}}: \simfunction(v_p, v_t) \leq t_{\simsubscript}\}\).

\paragraph{Continuous Metric}
For the continuous metric, only the predicted node with the highest similarity to each ground truth node above $t_{\simsubscript}$ is a true positive, i.e.,
\(
\{ v_p \in V_{i_{\predsubscript}} \,|\, \exists v_t \in V_{i_{\truesubscript}} : \simfunction(v_p, v_t) > t_{\simsubscript} \land \simfunction(v_p, v_t) = \max_{v_p' \in V_{i_{\predsubscript}}} \simfunction(v_p', v_t) \}.
\)
This stricter metric penalises multiple predictions for one ground truth node. 
We use this as the main metric, as it accounts for surface-form variations without allowing significantly different structures -- an approach that, through qualitative analysis, proved to reward the best ontologies in terms of downstream usability.
We match the hierarchy top-down: domains first, then slots, values, intents, and actions.
Only slots of matched domains and values of matched slots are considered.

\paragraph{Graph F1} 
On Wikipedia and arXiv, comparison results are from \citet{lo2024ontologylearning} using their evaluation scripts.
We report their literal, fuzzy, continuous, and graph F1 metrics.
Graph~F1 measures structural similarity between a predicted graph $G'$ and a gold graph $G$.
Each node is embedded using a 2-layer graph convolution, and nodes are matched by cosine similarity.
Let $s_{\text{graph}}$ be the total similarity of the best node matching.
Graph precision and recall are defined as:
\(
\text{Graph-Precision} = \frac{s_{\text{graph}}}{|V'|},
\text{Graph-Recall} = \frac{s_{\text{graph}}}{|V|},
\)
where $V'$ and $V$ are the node sets of $G'$ and $G$, respectively.
For more details, see \citet{lo2024ontologylearning}.


\subsection{Baselines}
\label{subsec:exp:baselines}


As DORE and GenDSI do not predict user intents or system actions, evaluation is limited to domains, slots, and values in \Cref{tab:sota_comparison}.

\paragraph{DORE \citep{vukovic-etal-2024-dialogue}} is defined as dialogue ontology relation extraction and fine-tunes Gemma-2B Instruct \citep{team2024gemma} to predict dialogue-level ontology relations between terms in the prompt.
Transfer learning is enhanced through constrained chain-of-thought decoding \citep{wang2024chain}, which selects the final response according to confidence across sampled responses while restricting decoding to known terms.
\Cref{tab:sota_comparison} shows the best-performing DORE models, including fine-tuned models and transfer models.

\begin{table*}[ht!]
    \centering
    \scriptsize
    \begin{tabular}{lccccccc}
        \toprule
        Approach & Domains & Slots & Values & Intents & Actions & Average & Supervised \\
        \midrule

        \multicolumn{2}{c}{\textit{MultiWOZ}} & \multicolumn{6}{c}{} \\
        \midrule


        {\TeQoDO} (ours) & \textbf{60.04} & \textbf{57.25} & 70.65 & 56.21 & 82.11 & \textbf{65.25} & \ding{55} \\
        
        DORE fine-tuned on MWOZ & 15.53 & 20.39 & \textbf{84.01} & - & - & 39.98 & \ding{51} \\

        DORE fine-tuned on SGD & 15.24 & 2.94 & 22.14 & - & - & 13.44 & \ding{55} \\

        GenDSI & 29.79 & 25.64 & 33.94 & - & - & 29.15 & \ding{55} \\

        \midrule
        \multicolumn{2}{c}{\textit{SGD}} & \multicolumn{6}{c}{} \\
        \midrule


        {\TeQoDO} (ours) & \textbf{72.19} & \textbf{43.70} & 48.57 & 76.48 & 67.28 & \textbf{61.64} & \ding{55}  \\

        DORE fine-tuned on MWOZ & 12.84 & 16.35 & 60.04 & - & - & 29.74 & \ding{55} \\

        DORE fine-tuned on SGD & 17.27 & 11.94 & \textbf{85.05} & - & - & 38.05 & \ding{51} \\

        GenDSI  & 27.07 & 29.08 & 41.28 & - & - & 32.47 & \ding{55} \\

        \bottomrule
    \end{tabular}
    \caption{
        Ontology construction comparison with continuous metric F1. 
        DORE \citep{vukovic-etal-2024-dialogue} and GenDSI \citep{finch-etal-2024-transforming} do not predict the intents and actions; hence, their average is only over domains, slots, and values.   
        We re-run both baselines using their published code bases.
        \label{tab:sota_comparison}
    }
    \vspace{-15pt}
\end{table*}

\paragraph{GenDSI \citep{finch-etal-2024-transforming}} is the generative dialogue state inference approach.
T5-3B \citep{raffel2020t5} is fine-tuned on a large synthetic TOD dataset \citep{finch-choi-2024-diverse} to generate dialogue-state updates from user–system turns.
Slot–value updates are clustered, and the most frequent slot name in each cluster is selected as the representative.
Domains are extracted from slot names by taking their first word to construct the ontology hierarchy for evaluation and comparison.
Predictions lacking domain names are discarded.

\paragraph{OLLM \citep{lo2024ontologylearning}}
is a Mistral 7B model \cite{jiang2023mistral7b} fine-tuned on Wikipedia and arXiv ontologies using a custom masking loss to make article-level ontology relation predictions.
These predictions are aggregated with frequency-based rules to form the final ontology.

\paragraph{Hearst patterns \cite{roller-etal-2018-hearst}} use hand-crafted lexico-syntactic rules, e.g.\ ``mammals such as humans'', to predict \emph{is-a} relations on Wikipedia and arXiv.
Additionally, \citet{lo2024ontologylearning} apply the smoothed relation matrix to weigh relations among the gold concepts.

\paragraph{REBEL \cite{huguet-cabot-navigli-2021-rebel-relation}} treats relation extraction as translation and trains a language model to generate relations on Wikipedia and arXiv.
\citet{lo2024ontologylearning} keep only relations of type ``subclass of'', ``instance of'', ``member of'', and ``part of'' and apply low-rank smoothing.

\subsection{{\TeQoDO}}
\label{subsec:exp:teqodo-model}
We use GPT-4o-mini (``gpt-4o-mini-2024-07-18'') \cite{gpt4o_blog_openai_2024} for all experiments.
The model performs well on HumanEval \cite{du2021humaneval}, which includes SQL coding tasks.
Full pipeline prompts are in \Cref{appendix:sec:prompt}.
SQLite (\Cref{footnote:sqlite}) is used via Python’s sqlite3.\footnote{\href{https://docs.python.org/3/library/sqlite3.html}{python sqlite3}}
Final prompts were manually crafted according to best practices\footnote{\label{footnote:openai_best_practices}\href{https://help.openai.com/en/articles/6654000-best-practices-for-prompt-engineering-with-the-openai-api}{OpenAI Prompting Best Practices}} and refined with ChatGPT (\Cref{subsec:experiments:promptphrasing}).

In the \emph{column value example} set-up, we sample values for each table column to expose the LLM to concrete instances for more consistency.
In the \emph{similarity matching} set-up, we compute semantic similarity using the ``all-MiniLM-L6-v2'' sentence-transformers model \cite{reimers-gurevych-2019-sentence} and a similarity threshold of $0.436$ (\Cref{footnote:threshold}).
For each query, we return up to 5 similar concepts that exceed this threshold, which is more versatile than using the \texttt{LIKE} operator from SQL for instance.

For general ontology data, we update the prompt to account for differing hierarchy levels.
We prompt the model to generate a DB capturing categories with parent IDs to represent more than three hierarchy levels.
A possible SQL representation uses columns for the category name and parent relation.
This follows the prompt from \citet{lo2024ontologylearning}.
We aggregate the ontology graph using category names and parent IDs to build hierarchical relations.
This ontology graph can be used directly in the \citet{lo2024ontologylearning} evaluation script.
On the arXiv dataset, we cluster top-level categories to account for high-level labels.

\subsection{Task-oriented Ontology Construction}
\label{subsec:exp:ontology_construction}

\paragraph{SOTA Comparison}
In \Cref{tab:sota_comparison} we see the continuous F1 evaluation of {\TeQoDO} compared to baselines.
It shows that DORE overfits to domains and slots in the training set but outperforms {\TeQoDO} on value-level and literal metrics.
This stems from supervised fine-tuning, enabling DORE to learn exact value phrasing.
GenDSI surpasses DORE on domains and slots but is outperformed by {\TeQoDO} on all metrics.
GenDSI’s lack of explicit domain prediction causes slot predictions to be dropped if domains are not the first word in slot names.
These results show that {\TeQoDO} predicts higher-level hierarchical concepts far better than the other methods.

\begin{table*}[ht]
    \centering
    \scriptsize
    \begin{tabular}{@{\hspace{0pt}}l@{\hspace{2pt}}@{\hspace{2pt}}ccc@{\hspace{2pt}}|@{\hspace{2pt}}ccc@{\hspace{2pt}}|@{\hspace{2pt}}ccc@{\hspace{0pt}}}
        \toprule
        Approach & \multicolumn{3}{c}{Literal} & \multicolumn{3}{c}{Fuzzy} & \multicolumn{3}{c}{Continuous} \\
        \midrule
        & F1 & Precision & Recall & F1 & Precision & Recall & F1 & Precision & Recall \\ 
        \midrule

        \multicolumn{4}{c}{\textit{MultiWOZ}} & \multicolumn{6}{c}{} \\
        \midrule

        Direct Update & \(2.8_{\pm 0.2}\) & \(2.5_{\pm 0.3}\) & \(16.1_{\pm 0.8}\) & \(37.7_{\pm 3.2}\) & \(32.2_{\pm 2.3}\) & \(88.1_{\pm 0.5}\) & \(19.2_{\pm 1.0}\) & \(19.9_{\pm 0.4}\) & \(58.5_{\pm 9.5}\) \\

        \midrule

        Query Update & \(13.1_{\pm 2.4}\) & \(14.7_{\pm 2.2}\) & \(13.6_{\pm 2.7}\) & \(62.3_{\pm 9.2}\) & \(60.7_{\pm 7.9}\) & \(77.7_{\pm 1.6}\) & \(50.1_{\pm 6.7}\) & \(45.0_{\pm 6.3}\) & \(73.4_{\pm 6.2}\) \\

        + DST Step & \(11.5_{\pm 3.5}\) & \(14.5_{\pm 4.5}\) & \(12.2_{\pm 4.0}\) & \(70.4_{\pm 4.7}\) & \(70.3_{\pm 6.2}\) & \(75.5_{\pm 3.8}\) & \(58.8_{\pm 4.3}\) & \(54.5_{\pm 4.9}\) & \(74.8_{\pm 3.8}\) \\

        + DST, Sim. & \(11.1_{\pm 3.1}\) & \(12.5_{\pm 4.4}\) & \(12.7_{\pm 3.7}\) & \(71.3_{\pm 3.1}\) & \(69.4_{\pm 4.8}\) & \(77.7_{\pm 2.9}\) & \(\mathbf{60.7}_{\pm 3.7}\) & \(55.0_{\pm 4.9}\) & \(77.1_{\pm 2.9}\) \\

        + DST, Ex.
        & \(12.1_{\pm 1.8}\) & \(16.4_{\pm 2.5}\) & \(12.4_{\pm 2.3}\) & \(\mathbf{74.7}_{\pm 4.0}\) & \(74.3_{\pm 5.9}\) & \(79.8_{\pm 2.2}\) & \(\mathbf{65.2}_{\pm 5.5}\) & \(61.2_{\pm 6.5}\) & \(78.5_{\pm 2.4}\) \\

        + DST, Ex., Success & \(12.3_{\pm 4.3}\) & \(16.8_{\pm 7.8}\) & \(12.9_{\pm 3.4}\) & \(73.3_{\pm 4.6}\) & \(75.3_{\pm 5.9}\) & \(76.2_{\pm 5.0}\) & \(\mathbf{65.2}_{\pm 4.7}\) & \(64.6_{\pm 7.0}\) & \(75.3_{\pm 4.7}\) \\

        \midrule
        \multicolumn{4}{c}{\textit{SGD}} & \multicolumn{6}{c}{} \\
        \midrule

        Direct Update
        & \(2.5_{\pm 0.2}\) & \(2.1_{\pm 0.3}\) & \(17.4_{\pm 0.2}\) & \(42.6_{\pm 7.0}\) & \(34.9_{\pm 7.5}\) & \(91.2_{\pm 0.7}\) & \(18.6_{\pm 2.3}\) & \(18.6_{\pm 1.6}\) & \(66.7_{\pm 13.0}\) \\

        \midrule

        Query Update
        & \(7.0_{\pm 2.7}\) & \(13.0_{\pm 5.5}\) & \(6.5_{\pm 3.4}\) & \(53.8_{\pm 24.2}\) & \(76.9_{\pm 20.3}\) & \(54.5_{\pm 28.1}\) & \(44.6_{\pm 21.1}\) & \(64.0_{\pm 19.7}\) & \(49.2_{\pm 26.6}\) \\

        + DST Step & \(10.0_{\pm 1.9}\) & \(15.7_{\pm 4.4}\) & \(8.9_{\pm 1.8}\) & \(75.0_{\pm 2.1}\) & \(89.1_{\pm 8.9}\) & \(69.1_{\pm 7.3}\) & \(61.4_{\pm 9.1}\) & \(72.2_{\pm 16.3}\) & \(61.9_{\pm 10.3}\) \\

        + DST, Sim. & \(7.0_{\pm 2.6}\) & \(11.8_{\pm 5.7}\) & \(6.5_{\pm 1.9}\) & \(70.1_{\pm 5.1}\) & \(85.5_{\pm 9.2}\) & \(65.2_{\pm 10.7}\) & \(53.4_{\pm 7.4}\) & \(65.6_{\pm 14.4}\) & \(55.1_{\pm 12.4}\) \\

        + DST, Ex.
        & \(9.0_{\pm 1.9}\) & \(14.7_{\pm 4.8}\) & \(8.2_{\pm 2.5}\) & \(73.5_{\pm 3.2}\) & \(90.3_{\pm 8.7}\) & \(66.1_{\pm 10.0}\) & \(60.8_{\pm 7.5}\) & \(74.3_{\pm 17.2}\) & \(59.5_{\pm 10.8}\) \\

        + DST, Ex., Success & \(8.3_{\pm 2.2}\) & \(14.7_{\pm 5.1}\) & \(7.0_{\pm 2.8}\) & \(70.2_{\pm 5.7}\) & \(93.4_{\pm 6.2}\) & \(60.5_{\pm 10.5}\) & \(\mathbf{61.6}_{\pm 5.4}\) & \(80.7_{\pm 12.5}\) & \(58.0_{\pm 10.7}\) \\

        \bottomrule
    \end{tabular}
    \caption{
        Ablation study for MultiWOZ and SGD test sets with 
        macro averages and standard deviations for 5 random dialogue orders over the five hierarchy classes.
        \textbf{Bold} F1 scores are significantly better than \emph{Query Update} (\(p < 0.05\)). 
        On SGD, we input batches of 10 dialogues to \emph{Query Update} approaches.
        }

        \label{tab:mwoz_plus_sgd_ontology_results}
    
    \vspace{-15pt} 
\end{table*}

\paragraph{Ablation}

\begin{table}[ht]
    \centering
    \scriptsize
    \begin{tabular}{lcc}
    \toprule
    \textbf{Approach} & \textbf{\# Tables} & \textbf{SQL Error Ratio \(\downarrow\)} \\
    \midrule
    \multicolumn{1}{c}{\textit{MultiWOZ (6 domains)}} & \multicolumn{2}{c}{} \\
    \midrule
    Direct Update & 168 & 28.72\% \\
    Query and Update & 14 & 37.31\% \\
    + DST & 12 & 30.45\% \\
    + DST, Sim & 16 & 23.19\% \\
    + DST, Ex. & \textbf{9} & 37.76\% \\
    + DST, Ex. Success & \textbf{9} & \textbf{4.71\%} \\

    \midrule
    \multicolumn{1}{c}{\textit{SGD (18 domains)}} & \multicolumn{2}{c}{} \\
    \midrule
    Direct Update & 432 & 30.27\% \\
    Query and Update & 92 & 32.50\% \\
    + DST & 69 & 32.81\% \\
    + DST, Sim & 88 & 17.71\% \\
    + DST, Ex. & 67 & 22.92\% \\
    + DST, Ex. Success & \textbf{37} & \textbf{5.22\%} \\

    \bottomrule
    \end{tabular}
    \caption{Comparison of Approaches by Number of Tables and SQL Error Ratio in update queries.}
    \label{tab:numtable_and_sqlerrors}
    \vspace{-17pt}  
\end{table}

\Cref{tab:mwoz_plus_sgd_ontology_results} shows the ablation study results, while \Cref{appendix:sec:per_class_results} details results for domains, slots, values, intents, and actions.
In the table, \emph{direct update} means generating a database update directly without first querying the existing database.
\emph{Query update} means generating an update only after querying the DB.
\emph{DST Step} indicates that the approach includes a form of DST, where the model predicts the structured state before forming the update.
\emph{Success} indicates that the prompt explicitly includes the user's goal or success condition.
Sim. denotes \emph{similarity matching}, where DB query results are expanded based on semantic similarity to increase the chance of results regardless of phrasing.
Ex. refers to \emph{column value examples}, where the column-information query result includes column–value pairs to guide update generation (See \Cref{subsec:TeQoDO:method,subsec:exp:teqodo-model} for details).

The discrepancy between literal and similarity-based metrics aligns with the findings of \citet{lo2024ontologylearning}, literal evaluation fails to capture surface form variations.
For example, a domain named ``hotel\_bookings'' is a false positive in literal evaluation despite being a reasonable prediction for the domain ``hotel''.
Precision on the continuous metric is lower than fuzzy, because only one prediction can match each ground truth concept.

\begin{table}[t]
    \setlength{\tabcolsep}{3pt}
    \centering
    \footnotesize
    \begin{tabular}{@{}lccccc|c@{}}
      \toprule
        \multirow{2}{*}{Ontology} & \multicolumn{5}{c}{Domains} \\ 
        & hotel & rest. & attr. & train & taxi & avg. \\
    \midrule
      Ground truth Ontology & \textbf{41.3} & \textbf{25.2} & 24.9 & \textbf{30.9} & 28.3 & 30.1 \\
      {\TeQoDO} Ontology & 26.7 & 23.9 & \textbf{44.8} & 29.1 & \textbf{33.2} & \textbf{31.5} \\
      \bottomrule
    \end{tabular}
    \caption{
        Zero-shot DST results for TripPy-R \citep{heck-etal-2022-robust} using ground truth vs {\TeQoDO} ontology on MultiWOZ 2.1 in JGA per domain.
        \label{tab:trippyr-zero-shot-groundtruth-vs-induced}
    }
    \vspace{-17pt}
\end{table}

Using the \emph{query and update pipeline} improves all scores on both datasets, already surpassing DORE in the more similarity-based metrics that are semantically more expressive according to \citet{lo2024ontologylearning}.
This is expected, as allowing the model to query existing tables leads to better alignment and more consistent update queries.
Without querying first, the model creates new table names for every dialogue, reducing precision.
Recall is higher in the \emph{direct update} approach due to more tables, but this substantially lowers precision.

\begin{table*}[t]
    \centering
    \scriptsize
    \begin{tabular}{lrrrrr}
        \toprule
        Approach & Literal F1 & Fuzzy F1 & Continuous F1 & Graph F1 & Unsupervised \\

        \midrule
        \multicolumn{2}{c}{\textit{Wikipedia}} & \multicolumn{4}{c}{} \\
        
        \midrule
        
        {\TeQoDO} & 0.03 & \textit{64.94} & 34.76 & \textit{64.15} & \ding{51} \\
        
        \midrule

        Hearst Patterns \citep{roller-etal-2018-hearst} & 0.30 & 53.80 & 35.00 & 54.40 & \ding{51} \\ 

        REBEL \citep{huguet-cabot-navigli-2021-rebel-relation} & \textit{0.40} & 62.40 & \textit{35.60} & 7.20 & \ding{55}  \\

        OLLM \citep{lo2024ontologylearning}  & \textbf{9.30} & \textbf{91.50} & \textbf{50.00} &  \textbf{64.40} & \ding{55} \\

        \midrule
        \multicolumn{2}{c}{\textit{arXiv}} & \multicolumn{4}{c}{} \\

        \midrule

        {\TeQoDO} + Clustering & 0.00 & \textit{33.95} & \textit{34.15} & \textbf{89.89} & \ding{51} \\

        \midrule

        Hearst Patterns \citep{roller-etal-2018-hearst} & 0.00 & 0.00 & 15.10 & 55.30 & \ding{51} \\ 

        REBEL \citep{huguet-cabot-navigli-2021-rebel-relation} & 0.00 & 6.00 & 28.10 & 54.60 & \ding{55}  \\ 

        OLLM \citep{lo2024ontologylearning}  & \textbf{4.00} & \textbf{57.00} & \textbf{35.70} & \textit{63.30} & \ding{55}  \\ 
        
        \bottomrule
    \end{tabular}
    \caption{Ontology construction results for Wikipedia and arXiv test sets from \cite{lo2024ontologylearning}. 
    We use the metrics from their code base and the Hearst, REBEL, and OLLM results from their work.
    In \textbf{bold} the best F1 score for each column is highlighted and the second highest in \textit{italics}.}
    \label{tab:arXiv_plus_wikipedia_ontology_results}
    \vspace{-15pt}
\end{table*}

On both datasets, incorporating concepts from modular TOD systems improves performance and reduces variance.
The \emph{DST step} alone does not significantly outperform the query and update baseline.
On MultiWOZ, \emph{similarity matching}, \emph{column value examples}, and \emph{success} yield significantly better performance.
On SGD, using \emph{success} together with \emph{column value examples} significantly improves the continuous metric.
Note that the combination of \emph{similarity matching} and \emph{success} yields no improvements, and is therefore omitted for brevity.
The concepts from modular TOD systems narrow the gap between fuzzy and continuous metrics, indicating more consistent DB updates.
Furthermore, the influence of dialogue order is diminished, as seen by the drop in variance. 

Comparing the \emph{direct update} baseline to the \emph{query and update} pipeline, the number of tables is significantly reduced.
As seen in \Cref{tab:numtable_and_sqlerrors}, the direct update baseline yields 168 tables for MultiWOZ, while the pipeline results in only 14, much closer to the ground truth of 6 domains.
Pipeline variants using \emph{DST} show similar table counts, which are further reduced by incorporating \emph{success}, closest to the 18 ground truth domains.
This reduction is even more pronounced on the SGD dataset.
The direct update baseline produces many redundant tables per domain, e.g., over 15 train-related domains.
In contrast, the pipeline typically results in one \texttt{train\_bookings} table.

Incorporating dialogue success into the DB update prompt significantly reduces erroneous SQL query ratios, achieving single-digit error rates on both datasets.
This suggests that applying concepts from modular TOD systems improves the model’s handling of the DB and SQL update queries, as most errors arise from incorrect column names, indicating poor schema handling.

\subsection{Downstream Application: DST}
\label{subsec:exp:dst_application}

To show downstream usability, we apply the {\TeQoDO}-induced ontology in a specialised dialogue state tracking model.
We replicate the zero-shot leave-one-domain-out setup from \citet{heck-etal-2022-robust}.
We train the TripPy-R model with one domain held out on the human-made ground truth ontology and do inference on that domain using the {\TeQoDO}-induced ontology instead of ground truth.
We use joint goal accuracy (JGA) as metric.

The mapping uses continuous evaluation, ensuring that at most one domain–slot pair is assigned to each ground truth slot.
During DST training, human-labelled slot names are replaced with predicted slots that match them.
If no predicted slot reaches the similarity threshold for a ground truth slot, that slot is excluded from training, and the model can never predict it, so its DST output remains empty.
For DST evaluation, predicted slots are mapped back to the ground truth slots.

See \Cref{tab:trippyr-zero-shot-groundtruth-vs-induced} for results comparing inference using the ground truth and {\TeQoDO}-induced ontology.
TripPy-R performs similarly with the induced ontology across all domains except ``hotel''.
In this case, the slots \textit{hotel-book people} and \textit{hotel-book stay} are missing due to unmapped slot predictions.
The ``restaurant'' and ``train'' domains also lack some slots, leading to lower performance.
In the ``taxi'' and ``attraction'' domains, all slots are mapped, and performance surpasses that with ground truth slots.
This may result from more informative slot names in the induced ontology, e.g. \textit{taxi\_bookings-pickup\_location} instead of ground truth \textit{taxi-departure}.
This shows that {\TeQoDO} can induce a useful ontology for downstream DST.

\begin{table*}[ht!]
    \centering
    \scriptsize
    \begin{tabular}{p{7.3cm}p{7.6cm}}
        \toprule

        Query Update & Query Update + DST \\
        
        \midrule

        \texttt{\textcolor{red}{INSERT INTO} hotel\_bookings (hotel\_name, location, price\_category, star\_rating) VALUES (`Alexander Bed and Breakfast', `Centre', `cheap', 4);} & \texttt{\textcolor{blue}{UPDATE hotel\_details SET} address = `56 saint barnabas road', phone\_number = `01223525725' WHERE name = `Alexander Bed and Breakfast' AND address IS NULL AND phone\_number IS NULL;} \\

        \midrule
        
        Query Update & Query Update + DST Similarity Matching \\
        \midrule
        \texttt{SELECT name, location, free\_wifi FROM guesthouses WHERE name = `Allenbell';} \newline $\rightarrow$ \textcolor{red}{No Result} 
        & 
        \texttt{SELECT name, location, free\_wifi FROM guesthouses WHERE name = `The Allenbell';} \newline $\rightarrow$ Result: \texttt{\textcolor{blue}{[(`The Allenbell', `east', 0)]}}  \\

        \midrule
        
        Query Update & Query Update + DST Column Value Examples \\

        \midrule

        \texttt{INSERT INTO intents (intent) VALUES 
        \textcolor{red}{('find\_swimming\_in\_east')};
        }& \texttt{INSERT INTO intents (intent\_name) VALUES \textcolor{blue}{('find\_pool')};} \\

        \midrule
        
        Query Update & Query Update + DST Value Examples and Success \\

        \midrule

        \texttt{INSERT INTO actions (action) VALUES 
        ('provide\_information'),
        ('recommend');}
        & \texttt{
        INSERT INTO system\_actions (action) VALUES \textcolor{blue}{('ask\_clarification')};} \\

        \bottomrule
    \end{tabular}
    \caption{Qualitative comparison between the query and update baseline and proposed improvements.
    }
    \label{tab:query_update_qualitative_analysis}
    \vspace{-15pt}
\end{table*}

\subsection{General Ontology Construction} 
\label{subsec:exp:general_data}

For general ontologies, we left-truncate the prompt to fit the model’s context window.  
This can remove earlier instructions and intermediate results from the final-step context.  
Because Wikipedia contains thousands of tables, pre-filtering in {\TeQoDO}’s first step may be necessary, which we leave for future work.

On arXiv, we evaluate only the first two hierarchy levels, as the ground truth contains only two.  
Given the dataset’s size and the small target ontology, directly applying {\TeQoDO} yields too many predicted tables.  
We therefore cluster table names and select representative tables by their proximity to each centroid.  
Unlike OLLM’s frequency-based aggregation, this removes the need for hand-crafted rules.  
We use \(k\)-means \cite{macqueen1967some} and choose \(k\) via the silhouette score \cite{ROUSSEEUW198753}, with \(k\) ranging from 5 to 20.  
To speed up inference, we input batches of $5$ articles for arXiv and $200$ for Wikipedia.

The challenges for {\TeQoDO} on general ontologies are dataset size and differing hierarchy depths compared to TOD ontologies.  
Hence, such ontologies are stored in SQL with category names and parent IDs for deeper hierarchies.  
These tables translate directly into an ontology graph by linking categories according to their parent IDs.  
The resulting graph is evaluated against the ground-truth using \citet{lo2024ontologylearning} code.  

\paragraph{Results}

See \Cref{tab:arXiv_plus_wikipedia_ontology_results} for {\TeQoDO} compared to models from \citet{lo2024ontologylearning} (see \Cref{appendix:subsec:general_examples} for examples).
On Wikipedia, {\TeQoDO} does not outperform OLLM on most metrics, likely due to OLLM's extensive supervised training.
However, {\TeQoDO} surpasses REBEL in fuzzy and graph F1, where it performs comparably to OLLM.
On the continuous metric, {\TeQoDO} performs on par with Hearst patterns and REBEL.

\begin{table*}[t]
\centering
\scriptsize
\begin{tabular}{lcccc|c}
    \toprule
    Approach & LLaMA 4 Scout (109B/17B) & Qwen3 (235B/22B) & Llama 3.1 (8B) & GPT-OSS (20B/3.6B) & GPT-4o-mini \\
    \midrule
    \multicolumn{2}{c}{\textit{Spider multi-turn text-to-SQL exact match}} &
    \multicolumn{4}{c}{} \\
    Text-to-SQL & 27.1 & 29.0 & 25.9 & 22.4 & 40.2 \\
    
    \midrule
    \multicolumn{2}{c}{\textit{MultiWOZ Continuous F1}} &
    \multicolumn{4}{c}{} \\
    Direct Updates & $21.4_{\pm 5.5}$ & $32.6_{\pm 14.5}$ & $22.8_{\pm 6.0}$ & $21.9_{\pm 3.0}$ & $19.2_{\pm 1.0}$ \\
    DB Query Update & $28.8_{\pm 11.9}$ & $32.8_{\pm 6.3}$ & $22.5_{\pm 3.0}$ & $43.8_{\pm 6.2}$ & $50.1_{\pm 6.7}$ \\
    + DST, Ex., Succ. & $33.4_{\pm 5.4}$ & $51.9_{\pm 7.1}$ & $25.3_{\pm 3.2}$ & $40.2_{\pm 5.6}$ & $65.2_{\pm 4.7}$  \\
    \midrule
    \multicolumn{2}{c}{\textit{SGD Continuous F1}} &
    \multicolumn{4}{c}{} \\
    Direct Updates & $39.6_{\pm 4.9}$ & $30.6_{\pm 1.1}$ & $18.5_{\pm 1.3}$ & $47.8_{\pm 11.6}$ & $18.6_{\pm 2.3}$ \\
    DB Query Update & $54.3_{\pm 7.9}$ & $60.3_{\pm 7.3}$ & $41.4_{\pm 8.3}$ & $68.3_{\pm 5.0}$ & $44.6_{\pm 21.1}$ \\
    + DST, Ex., Succ. & $63.2_{\pm 4.0}$ & $67.7_{\pm 3.9}$ & $55.7_{\pm 4.1}$ & $57.8_{\pm 4.9}$ & $61.6_{\pm 5.4}$  \\
    \bottomrule
\end{tabular}
\caption{Continuous F1 scores and standard deviations of open models of different sizes and GPT-4o-mini with 5 random dialogue-order seeds.
For MoE models, we indicate both the total parameter count and the active parameters per forward pass.
On SGD, we use batches of 10 dialogues to speed up inference.
Additionally, we show the multi-turn exact match performance on the Spider dataset \cite{yu-etal-2018-spider}.
}
\label{tab:f1_open_models_mwoz_sgd}
\vspace{-15pt}
\end{table*}

By clustering tables, obtain competitive results on arXiv.  
{\TeQoDO} achieves the highest graph F1 and second-best continuous F1 on arXiv, outperforming fine-tuned and few-shot models and matching supervised OLLM.  
This indicates that {\TeQoDO}’s induced ontologies closely match the ground truth structure.  
In fuzzy and continuous F1, {\TeQoDO} surpasses Hearst patterns and REBEL.  
Lower literal F1 scores follow from the absence of supervision in {\TeQoDO}.  
It adapts to ontologies beyond TOD, though arXiv requires specific adjustments.  
The differing hierarchy depth poses challenges, as SQL fits the three-level TOD structure best.  
We argue that arXiv’s high-level ontology, with only 61 nodes for thousands of abstracts, underrepresents its content richness.

\subsection{TOD Ontology Qualitative Analysis}
\label{subsec:experiments:qualitative}


See \Cref{tab:query_update_qualitative_analysis} for a qualitative comparison of update queries on MultiWOZ test dialogues, detailed in \Cref{appendix:sec:example_dialogues}.
In the first example, we illustrate the impact of adding the \emph{DST step}, which clarifies the distinction between new dialogue information and existing DB content.
Without DST, the model inserts a new entry for ``Alexander Bed and Breakfast''.
With DST, it updates the existing DB entry by adding only the missing information.

In the second example, ``Allenbell'' appears in the dialogue, while ``The Allenbell'' is stored from a previous dialogue.
Without similarity matching, the model queries ``Allenbell'', yielding no result due to the missing ``The''.
Similarity matching retrieves ``The Allenbell'' from the DB.
The model chooses which results to include, as less related concepts may also be retrieved.

\emph{Column value examples} help with more general intents and actions, e.g., find\_pool'' instead of ``find\_swimming\_in\_east''.
Mentioning dialogue success encourages system actions like ``ask\_clarification'' that support user goals. See \Cref{appendix:sec:ontology_excerpt} for predicted ontology excerpts. 


To estimate the frequency of the observations, we sample $50$ dialogues from the MultiWOZ test set and relate qualitative observations to quantitative improvement.
For example, if a qualitative effect on user intents appears in \(2/50\) dialogues, we expect the corresponding metric to improve by roughly \(4\%\).


In \(9/50\) cases \((18\%)\), query updates + DST \emph{updates} rather than inserting, matching the macro F1 rising from \(50.1\) to \(58.8\), \(15\%\) relative.
In \(2/50\) cases \((4\%)\), query updates with DST and similarity matching yield an \emph{update} based on the expanded DB results.
This better aligns user intents and system actions, leading to a \(3\%\) macro gain over DST alone.
In \(8/50\) cases \((16\%)\), adding DST and column-value examples produces more general user-intent names.
This corresponds to a relative macro F1 gain of about \(20\%\).
Over DST alone, we observe a \(4\%\) gain.
In \(3/50\) cases \((6\%)\), additionally adding success produces more user-focussed system actions.
Here, intent F1 improves by \(20\%\) and action F1 by around \(10\%\) compared to DST plus examples.
This shows that the observed behaviour is in line with the ablation.

\subsection{Open-source Model Inference}

To show the generalisation of TeQoDO to other LLMs, we use it with four open-source models of varying sizes using Vertex AI\footnote{\url{https://cloud.google.com/vertex-ai}}.
We use LLaMa-4-scout \cite{MetaAI2025_Llama4Blog}, a multimodal Mixture of Experts \cite[MoE;][]{Jacobs1991AdaptiveMixtures,Shazeer2017OutrageouslyLargeMoE} with 109B total and 17B active parameters per forward pass.
We also use Qwen-3-235B \cite{qwen3technicalreport} with 22B active parameters and GPT-OSS-20B \cite{OpenAI2025_GPT-OSS} with 3.6B active parameters.
As a small non-MoE model, we use LLaMa-3.1-8B \cite{MetaAI2024_LLaMA31Blog}.

See \Cref{tab:f1_open_models_mwoz_sgd} for the results, where we additionally show the exact match text-to-SQL performance in multi-turn instructions, based on \citet{laban2025llms} sharded instruction set-up on the Spider text-to-SQL dataset \cite{yu-etal-2018-spider}.
All models generate useful SQL statements to build a database from the dialogues, showing TeQoDO's usability across LLMs.
The larger models, LLaMa-4-scout and Qwen-3, perform better with the query and update pipeline.
Including the DST step and success further enhances performance, approaching GPT-4o-mini on SGD, despite being less capable in text-to-SQL.

Interestingly, the improvement of LLaMa models is less pronounced on MultiWOZ.
Here, LLaMa models frequently generate \texttt{CREATE TABLE} queries even with the query-update pipeline.
The small LLaMa-3.1-8B model generates \texttt{CREATE TABLE} queries without \texttt{IF NOT EXISTS}, causing overprediction of tables and lower performance than other models.
However, on SGD, LLaMa-3.1-8B improves with TeQoDO, as the generated table names have less variation.

For GPT-OSS-20B, which has the fewest active parameters, the DST step and dialogue success in the full {\TeQoDO} pipeline do not improve performance.
We assume the small expert cannot model state and dialogue success while interacting with the SQL database.
This is reflected in lower domain and user intent performance when using the DST step and dialogue success.
Qualitatively, the model overengineers the database, generating more update queries and more specific domains and user intents, which decreases performance.

These results demonstrate the generalisability of {\TeQoDO} to open models as small as 8B parameters and an MoE model with 3.6B active parameters.
Moreover, while some text-to-SQL expertise of the LLM is necessary for this approach to work, {\TeQoDO} does compensate to a certain extent for LLMs that lack strong text-to-SQL abilities. 
The query-update pipeline and the use of DST and success benefit larger models more.

\subsection{Prompt phrasing Analysis} 
\label{subsec:experiments:promptphrasing}


\begin{figure}[t]
    \centering
    \includegraphics[width=\linewidth]{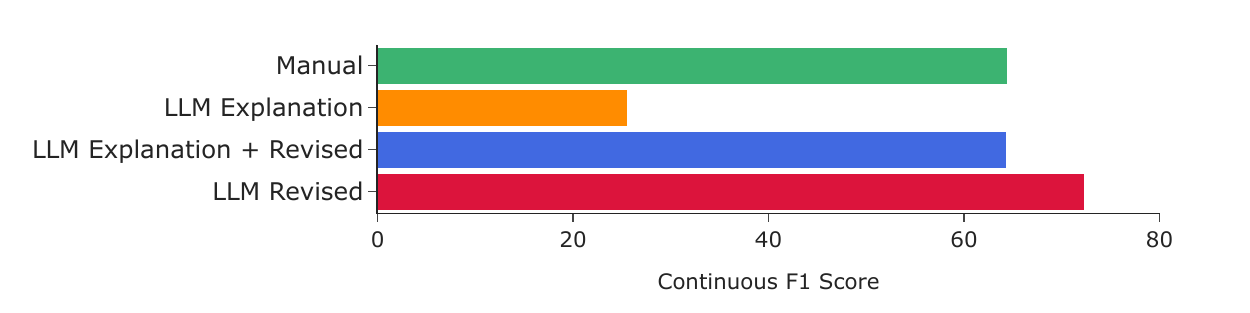}
    \caption{Continuous F1 of different Prompts on MultiWOZ test-set based on LLM explanation.}
    \label{fig:different_llm_prompt_figure}
    \vspace{-15pt}
\end{figure}

We begin with a manually written prompt following OpenAI best practices (\Cref{footnote:openai_best_practices}).
Next, we prompt GPT-4o to explain its understanding of this initial prompt.
Finally, the LLM revises the prompt using its explanation, facilitating task adherence for LLMs, similar to \citet{Kim2024ReEx}.

The explanation-revised prompt in \Cref{fig:different_llm_prompt_figure} improves MultiWOZ performance by around 10\% over the manual prompt.
This gain comes from increased clarity -- shorter sentences, bullet points, and direct language -- which yields more general database updates.
For instance, the revised prompt clearly separates desired and non-desired behaviours.
The explanation alone performs worse, as it lacks explicit instructions.

Mentioning success only at the end of the update prompt improves performance over not mentioning it, particularly for intents and actions.
If mentioned too early, the model ignores the user goal in the update.
This shows that performance depends on the phrasing of the prompt, a general issue for LLMs \citep{razavi2025promptsensitivity}, but TeQoDO still works with different prompts.

\subsection{Discussion and Future Work}
\label{subsec:res:discussion}

Our results show that SQL’s structured format improves ontology quality when a variety of LLMs with different levels of text-to-SQL abilities interact with a DB.
The SQL-enhanced model aligns values more accurately using \emph{semantic matching} or \emph{column value examples}.
Modular TOD system concepts help in distinguishing existing database information from new input.
{\TeQoDO} significantly outperforms the supervisedly fine-tuned DORE and GenDSI.
It also reduces sensitivity to dialogue order.
Future work will focus on enabling the model to restructure the DB after iterating over a dataset instead of clustering. 

Regarding training data contamination~\cite{deng-etal-2024-unveiling,samuel-etal-2025-towards}, the results of the \emph{direct update} approach show that the model has not memorised the TOD datasets, as it cannot recall the exact table names for each dialogue.

For larger or structurally different ontologies than TOD, our results show {\TeQoDO} can generalise, though SQL struggles to express multiple hierarchy levels.
We aim to scale {\TeQoDO} to diverse ontology structures by adapting the text-to-SQL component; however, its sequential nature limits parallelisation.
The large batch size for concatenated Wikipedia articles may affect performance, though evaluating this is computationally expensive.
Finally, our approach is sensitive to prompt phrasing, a common issue with large language models \cite{razavi2025promptsensitivity}.

Our evaluation captures all necessary information and is more fine-grained than existing methods by considering the hierarchy levels of TOD ontologies: domains, slots, values, system actions, and user intents.
Using an off-the-shelf similarity model with a fixed threshold may introduce bias, though qualitative analysis is aligned with the quantitative results.
In future work, we aim to incorporate human judgment into evaluation to reduce reliance on fixed thresholds.

\citet{dennett1987intentional} argues that task comprehension is not required for competent performance.
Inspired by this, we think that the LLMs’ competence in SQL generation can be used to extract task knowledge in the form of an ontology.
We see this work as an important step in distilling human-readable knowledge from text that can be used as a knowledge source for otherwise black-box LLMs. 
Extracting ontologies automatically with LLMs and analysing their structure may enhance their interpretability in downstream tasks.

\section{Conclusion}
\label{sec:conclusion}

We introduce {\TeQoDO}, a method that uses large language models to build ontologies from task-oriented data, exploiting the inherent SQL programming abilities of LLMs and incorporating modular TOD concepts.
We evaluate {\TeQoDO} on two widely used TOD datasets and find that it surpasses current SOTA.
We also show that {\TeQoDO} generalises to a variety of LLMs and to large ontologies with different structures.
Our results motivate further exploration of this approach for explainability. 

\section*{Acknowledgements}

We thank the reviewers and the action editor for their insightful comments and suggested revisions, which significantly improved the quality of this paper.
This work was funded by the European Research Council (ERC) under the Horizon 2020 research and innovation program (Grant No. STG2018 804636) and the Ministry of Culture and Science of North Rhine-Westphalia within the Lamarr Fellow Network.
Computational resources were provided by the Centre for Information and Media Technology at Heinrich Heine University Düsseldorf and Google Cloud.

\newpage
\vspace{12pt}

\bibliographystyle{acl_natbib}
\bibliography{tacl2021}

@inproceedings{feng-etal-2024-infusing,
    title = {{Infusing Emotions into Task-oriented Dialogue Systems: Understanding, Management, and Generation}},
    author = "Feng, Shutong  and
      Lin, Hsien-chin  and
      Geishauser, Christian  and
      Lubis, Nurul  and
      van Niekerk, Carel  and
      Heck, Michael  and
      Ruppik, Benjamin Matthias  and
      Vukovic, Renato  and
      Gasic, Milica",
    editor = "Kawahara, Tatsuya  and
      Demberg, Vera  and
      Ultes, Stefan  and
      Inoue, Koji  and
      Mehri, Shikib  and
      Howcroft, David  and
      Komatani, Kazunori",
    booktitle = "Proceedings of the 25th Annual Meeting of the Special Interest Group on Discourse and Dialogue",
    month = sep,
    year = "2024",
    address = "Kyoto, Japan",
    publisher = "Association for Computational Linguistics",
    url = "https://aclanthology.org/2024.sigdial-1.60/",
    doi = "10.18653/v1/2024.sigdial-1.60",
    pages = "699--717"
}

@inproceedings{vukovic-etal-2022-dialogue,
    title = {{Dialogue Term Extraction using Transfer Learning and Topological Data Analysis}},
    author = "Vukovic, Renato  and
      Heck, Michael  and
      Ruppik, Benjamin  and
      van Niekerk, Carel  and
      Zibrowius, Marcus  and
      Gasic, Milica",
    editor = "Lemon, Oliver  and
      Hakkani-Tur, Dilek  and
      Li, Junyi Jessy  and
      Ashrafzadeh, Arash  and
      Garcia, Daniel Hern{\'a}ndez  and
      Alikhani, Malihe  and
      Vandyke, David  and
      Du{\v{s}}ek, Ond{\v{r}}ej",
    booktitle = "Proceedings of the 23rd Annual Meeting of the Special Interest Group on Discourse and Dialogue",
    month = sep,
    year = "2022",
    address = "Edinburgh, UK",
    publisher = "Association for Computational Linguistics",
    url = "https://aclanthology.org/2022.sigdial-1.53/",
    doi = "10.18653/v1/2022.sigdial-1.53",
    pages = "564--581"
}

@inproceedings{vukovic-etal-2024-dialogue,
    title = {{Dialogue Ontology Relation Extraction via Constrained Chain-of-Thought Decoding}},
    author = "Vukovic, Renato  and
      Arps, David  and
      van Niekerk, Carel  and
      Ruppik, Benjamin Matthias  and
      Lin, Hsien-chin  and
      Heck, Michael  and
      Gasic, Milica",
    editor = "Kawahara, Tatsuya  and
      Demberg, Vera  and
      Ultes, Stefan  and
      Inoue, Koji  and
      Mehri, Shikib  and
      Howcroft, David  and
      Komatani, Kazunori",
    booktitle = "Proceedings of the 25th Annual Meeting of the Special Interest Group on Discourse and Dialogue",
    month = sep,
    year = "2024",
    address = "Kyoto, Japan",
    publisher = "Association for Computational Linguistics",
    url = "https://aclanthology.org/2024.sigdial-1.33/",
    doi = "10.18653/v1/2024.sigdial-1.33",
    pages = "370--384"
}

@inproceedings{lo2024ontologylearning,
 author = {Lo, Andy and Jiang, Albert Q and Li, Wenda and Jamnik, Mateja},
 booktitle = {Advances in Neural Information Processing Systems},
 publisher = {Curran Associates, Inc.},
 title = {{End-to-End Ontology Learning with Large Language Models}},
 volume = {38},
 year = {2024}
}

@inproceedings{li2023llmDBinterface,
author = {Li, Jinyang and Hui, Binyuan and Qu, Ge and Yang, Jiaxi and Li, Binhua and Li, Bowen and Wang, Bailin and Qin, Bowen and Geng, Ruiying and Huo, Nan and Zhou, Xuanhe and Ma, Chenhao and Li, Guoliang and Chang, Kevin C.C. and Huang, Fei and Cheng, Reynold and Li, Yongbin},
title = {{Can LLM Already Serve as A Database Interface? A Big Bench for Large-Scale Database Grounded Text-to-SQLs}},
year = {2024},
publisher = {Curran Associates Inc.},
address = {Red Hook, NY, USA},
booktitle = {Proceedings of the 37th International Conference on Neural Information Processing Systems},
articleno = {1835},
numpages = {28},
location = {New Orleans, LA, USA},
series = {NIPS '23}
}

@article{zhou_db-gpt_2024,
	title = {{DB}-{GPT}: {Large} {Language} {Model} {Meets} {Database}},
	volume = {9},
	issn = {2364-1541},
	url = {https://doi.org/10.1007/s41019-023-00235-6},
	doi = {10.1007/s41019-023-00235-6},
	number = {1},
	journal = {Data Science and Engineering},
	author = {Zhou, Xuanhe and Sun, Zhaoyan and Li, Guoliang},
	month = mar,
	year = {2024},
	pages = {102--111},
}

@inproceedings{rastogi2020towards,
  title={{Towards Scalable Multi-domain Conversational Agents: The Schema-guided Dialogue Dataset}},
  author={Rastogi, Abhinav and Zang, Xiaoxue and Sunkara, Srinivas and Gupta, Raghav and Khaitan, Pranav},
  booktitle={Proceedings of the AAAI Conference on Artificial Intelligence},
  volume={34},
  pages={8689--8696},
  year={2020}
}

@inproceedings{eric-etal-2020-multiwoz,
    title = "{{M}ulti{WOZ} 2.1: A Consolidated Multi-Domain Dialogue Dataset with State Corrections and State Tracking Baselines}",
    author = "Eric, Mihail  and
      Goel, Rahul  and
      Paul, Shachi  and
      Sethi, Abhishek  and
      Agarwal, Sanchit  and
      Gao, Shuyang  and
      Kumar, Adarsh  and
      Goyal, Anuj  and
      Ku, Peter  and
      Hakkani-Tur, Dilek",
    editor = "Calzolari, Nicoletta  and
      B{\'e}chet, Fr{\'e}d{\'e}ric  and
      Blache, Philippe  and
      Choukri, Khalid  and
      Cieri, Christopher  and
      Declerck, Thierry  and
      Goggi, Sara  and
      Isahara, Hitoshi  and
      Maegaard, Bente  and
      Mariani, Joseph  and
      Mazo, H{\'e}l{\`e}ne  and
      Moreno, Asuncion  and
      Odijk, Jan  and
      Piperidis, Stelios",
    booktitle = "Proceedings of the Twelfth Language Resources and Evaluation Conference",
    month = may,
    year = "2020",
    address = "Marseille, France",
    publisher = "European Language Resources Association",
    url = "https://aclanthology.org/2020.lrec-1.53/",
    pages = "422--428",
    language = "eng",
    ISBN = "979-10-95546-34-4"
}

@inproceedings{zhu-etal-2023-convlab-local,
    title = {{{C}onv{L}ab-3: A Flexible Dialogue System Toolkit Based on a Unified Data Format}},
    author = "Zhu, Qi  and
      Geishauser, Christian  and
      Lin, Hsien-chin  and
      van Niekerk, Carel  and
      Peng, Baolin  and
      Zhang, Zheng  and
      Feng, Shutong  and
      Heck, Michael  and
      Lubis, Nurul  and
      Wan, Dazhen  and
      Zhu, Xiaochen  and
      Gao, Jianfeng  and
      Gašić, Milica  and
      Huang, Minlie",
    editor = "Feng, Yansong  and
      Lefever, Els",
    booktitle = "Proceedings of the 2023 Conference on Empirical Methods in Natural Language Processing: System Demonstrations",
    month = dec,
    year = "2023",
    address = "Singapore",
    publisher = "Association for Computational Linguistics",
    url = "https://aclanthology.org/2023.emnlp-demo.9",
    doi = "10.18653/v1/2023.emnlp-demo.9",
    pages = "106--123",
}

@inproceedings{brown2020gpt3,
    author = {Brown, Tom and Mann, Benjamin and Ryder, Nick and Subbiah, Melanie and Kaplan, Jared D and Dhariwal, Prafulla and Neelakantan, Arvind and Shyam, Pranav and Sastry, Girish and Askell, Amanda and Agarwal, Sandhini and Herbert-Voss, Ariel and Krueger, Gretchen and Henighan, Tom and Child, Rewon and Ramesh, Aditya and Ziegler, Daniel and Wu, Jeffrey and Winter, Clemens and Hesse, Chris and Chen, Mark and Sigler, Eric and Litwin, Mateusz and Gray, Scott and Chess, Benjamin and Clark, Jack and Berner, Christopher and McCandlish, Sam and Radford, Alec and Sutskever, Ilya and Amodei, Dario},
    booktitle = {Advances in Neural Information Processing Systems},
    editor = {H. Larochelle and M. Ranzato and R. Hadsell and M.F. Balcan and H. Lin},
    pages = {1877--1901},
    publisher = {Curran Associates, Inc.},
    title = {{Language Models are Few-Shot Learners}},
    url = {https://proceedings.neurips.cc/paper_files/paper/2020/file/1457c0d6bfcb4967418bfb8ac142f64a-Paper.pdf},
    volume = {33},
    year = {2020}
}

@inproceedings{ouyang2022instructgpt,
    author = {Ouyang, Long and Wu, Jeffrey and Jiang, Xu and Almeida, Diogo and Wainwright, Carroll and Mishkin, Pamela and Zhang, Chong and Agarwal, Sandhini and Slama, Katarina and Ray, Alex and Schulman, John and Hilton, Jacob and Kelton, Fraser and Miller, Luke and Simens, Maddie and Askell, Amanda and Welinder, Peter and Christiano, Paul F and Leike, Jan and Lowe, Ryan},
    booktitle = {Advances in Neural Information Processing Systems},
    editor = {S. Koyejo and S. Mohamed and A. Agarwal and D. Belgrave and K. Cho and A. Oh},
    pages = {27730--27744},
    publisher = {Curran Associates, Inc.},
    title = {{Training Language Models to follow Instructions with Human Feedback}},
    url = {https://proceedings.neurips.cc/paper_files/paper/2022/file/b1efde53be364a73914f58805a001731-Paper-Conference.pdf},
    volume = {35},
    year = {2022}
}

@inproceedings{cifka-liutkus-2023-black,
    title = "{Black-box Language Model Explanation by Context Length Probing}",
    author = "C{\'i}fka, Ond{\v{r}}ej  and
      Liutkus, Antoine",
    editor = "Rogers, Anna  and
      Boyd-Graber, Jordan  and
      Okazaki, Naoaki",
    booktitle = "Proceedings of the 61st Annual Meeting of the Association for Computational Linguistics (Volume 2: Short Papers)",
    month = jul,
    year = "2023",
    address = "Toronto, Canada",
    publisher = "Association for Computational Linguistics",
    url = "https://aclanthology.org/2023.acl-short.92/",
    doi = "10.18653/v1/2023.acl-short.92",
    pages = "1067--1079"
}

@inproceedings{zhong2024knowledge,
    title={{Seeking Neural Nuggets: Knowledge Transfer in Large Language Models from a Parametric Perspective}},
    author={Zhong, Ming and An, Chenxin and Chen, Weizhu and Han, Jiawei and He, Pengcheng},
    booktitle = {The Twelfth International Conference on Learning Representations (ICLR)},
    year = 2024
}

@inproceedings{sahoo-etal-2024-comprehensive,
    title = {{A Comprehensive Survey of Hallucination in Large Language, Image, Video and Audio Foundation Models}},
    author = "Sahoo, Pranab  and
      Meharia, Prabhash  and
      Ghosh, Akash  and
      Saha, Sriparna  and
      Jain, Vinija  and
      Chadha, Aman",
    editor = "Al-Onaizan, Yaser  and
      Bansal, Mohit  and
      Chen, Yun-Nung",
    booktitle = "Findings of the Association for Computational Linguistics: EMNLP 2024",
    month = nov,
    year = "2024",
    address = "Miami, Florida, USA",
    publisher = "Association for Computational Linguistics",
    url = "https://aclanthology.org/2024.findings-emnlp.685/",
    doi = "10.18653/v1/2024.findings-emnlp.685",
    pages = "11709--11724"
}

@inproceedings{hudecek-dusek-2023-large,
    title = {{Are Large Language Models All You Need for Task-Oriented Dialogue?}},
    author = "Hude{\v{c}}ek, Vojt{\v{e}}ch  and
      Dusek, Ondrej",
    editor = "Stoyanchev, Svetlana  and
      Joty, Shafiq  and
      Schlangen, David  and
      Dusek, Ondrej  and
      Kennington, Casey  and
      Alikhani, Malihe",
    booktitle = "Proceedings of the 24th Annual Meeting of the Special Interest Group on Discourse and Dialogue",
    month = sep,
    year = "2023",
    address = "Prague, Czechia",
    publisher = "Association for Computational Linguistics",
    url = "https://aclanthology.org/2023.sigdial-1.21/",
    doi = "10.18653/v1/2023.sigdial-1.21",
    pages = "216--228"
}

@article{chen2022program,
  title = {{Program of Thoughts Prompting: Disentangling Computation from Reasoning for Numerical Reasoning Tasks}},
  author = {Wenhu Chen and Xueguang Ma and Xinyi Wang and William W. Cohen},
  journal={Transactions on Machine Learning Research},
  year = {2023},
}

@inproceedings{deng-etal-2022-recent,
    title = {{Recent Advances in Text-to-{SQL}: A Survey of What We Have and What We Expect}},
    author = "Deng, Naihao  and
      Chen, Yulong  and
      Zhang, Yue",
    editor = "Calzolari, Nicoletta  and
      Huang, Chu-Ren  and
      Kim, Hansaem  and
      Pustejovsky, James  and
      Wanner, Leo  and
      Choi, Key-Sun  and
      Ryu, Pum-Mo  and
      Chen, Hsin-Hsi  and
      Donatelli, Lucia  and
      Ji, Heng  and
      Kurohashi, Sadao  and
      Paggio, Patrizia  and
      Xue, Nianwen  and
      Kim, Seokhwan  and
      Hahm, Younggyun  and
      He, Zhong  and
      Lee, Tony Kyungil  and
      Santus, Enrico  and
      Bond, Francis  and
      Na, Seung-Hoon",
    booktitle = "Proceedings of the 29th International Conference on Computational Linguistics",
    month = oct,
    year = "2022",
    address = "Gyeongju, Republic of Korea",
    publisher = "International Committee on Computational Linguistics",
    url = "https://aclanthology.org/2022.coling-1.190/",
    pages = "2166--2187"
}

@misc{gpt4o_blog_openai_2024, 
    title={{GPT-4o mini: advancing cost-efficient intelligence}}, 
    url={https://openai.com/index/gpt-4o-mini-advancing-cost-efficient-intelligence/}, 
    journal={OpenAI Blog}, 
    author={OpenAI}, 
    year={2024}, 
    month={July},
    note={Accessed 2025-01-13}
}

@inproceedings{miller-1994-wordnet,
    title = "{{W}ord{N}et: A Lexical Database for {E}nglish}",
    author = "Miller, George A.",
    booktitle = "{H}uman {L}anguage {T}echnology: Proceedings of a Workshop held at {P}lainsboro, {N}ew {J}ersey, {M}arch 8-11, 1994",
    year = "1994",
    url = "https://aclanthology.org/H94-1111/"
}

@inproceedings{reimers-gurevych-2019-sentence,
    title = {{Sentence-{BERT}: Sentence Embeddings using {S}iamese {BERT}-Networks}},
    author = "Reimers, Nils  and
      Gurevych, Iryna",
    editor = "Inui, Kentaro  and
      Jiang, Jing  and
      Ng, Vincent  and
      Wan, Xiaojun",
    booktitle = "Proceedings of the 2019 Conference on Empirical Methods in Natural Language Processing and the 9th International Joint Conference on Natural Language Processing (EMNLP-IJCNLP)",
    month = nov,
    year = "2019",
    address = "Hong Kong, China",
    publisher = "Association for Computational Linguistics",
    url = "https://aclanthology.org/D19-1410/",
    doi = "10.18653/v1/D19-1410",
    pages = "3982--3992"
}

@article{team2024gemma,
  title={{Gemma 2: Improving Open Language Models at a Practical Size}},
  author={Team, Gemma and Riviere, Morgane and Pathak, Shreya and Sessa, Pier Giuseppe and Hardin, Cassidy and Bhupatiraju, Surya and Hussenot, L{\'e}onard and Mesnard, Thomas and Shahriari, Bobak and Ram{\'e}, Alexandre and others},
  journal={arXiv preprint arXiv:2408.00118},
  year={2024}
}

@inproceedings{yang-etal-2024-synthesizing,
    title = {{Synthesizing Text-to-{SQL} Data from Weak and Strong {LLM}s}},
    author = "Yang, Jiaxi  and
      Hui, Binyuan  and
      Yang, Min  and
      Yang, Jian  and
      Lin, Junyang  and
      Zhou, Chang",
    editor = "Ku, Lun-Wei  and
      Martins, Andre  and
      Srikumar, Vivek",
    booktitle = "Proceedings of the 62nd Annual Meeting of the Association for Computational Linguistics (Volume 1: Long Papers)",
    month = aug,
    year = "2024",
    address = "Bangkok, Thailand",
    publisher = "Association for Computational Linguistics",
    url = "https://aclanthology.org/2024.acl-long.425/",
    doi = "10.18653/v1/2024.acl-long.425",
    pages = "7864--7875"
}

@inproceedings{allemang2024ontologies,
author = {Allemang, Dean and Sequeda, Juan},
title = {{Increasing the Accuracy of LLM Question-Answering Systems with Ontologies}},
year = {2024},
isbn = {978-3-031-77846-9},
publisher = {Springer-Verlag},
address = {Berlin, Heidelberg},
url = {https://doi.org/10.1007/978-3-031-77847-6_18},
doi = {10.1007/978-3-031-77847-6_18},
booktitle = {The Semantic Web – ISWC 2024: 23rd International Semantic Web Conference, Baltimore, MD, USA, November 11–15, 2024, Proceedings, Part III},
pages = {324–339},
numpages = {16},
location = {Hanover, MD, USA}
}

@inproceedings{zhao-etal-2024-deciphering,
    title = {{Deciphering the Impact of Pretraining Data on Large Language Models through Machine Unlearning}},
    author = "Zhao, Yang  and
      Du, Li  and
      Ding, Xiao  and
      Xiong, Kai  and
      Sun, Zhouhao  and
      Jun, Shi  and
      Liu, Ting  and
      Qin, Bing",
    editor = "Ku, Lun-Wei  and
      Martins, Andre  and
      Srikumar, Vivek",
    booktitle = "Findings of the Association for Computational Linguistics: ACL 2024",
    month = aug,
    year = "2024",
    address = "Bangkok, Thailand",
    publisher = "Association for Computational Linguistics",
    url = "https://aclanthology.org/2024.findings-acl.559/",
    doi = "10.18653/v1/2024.findings-acl.559",
    pages = "9386--9406"
}

@inproceedings{Polleres2014sparql,
author="Polleres, Axel",
editor="Alhajj, Reda
and Rokne, Jon",
title={{SPARQL}},
bookTitle="Encyclopedia of Social Network Analysis and Mining",
year="2014",
publisher="Springer New York",
address="New York, NY",
pages="1960--1966",
isbn="978-1-4614-6170-8",
doi="10.1007/978-1-4614-6170-8_124",
url="https://doi.org/10.1007/978-1-4614-6170-8_124"
}

@inproceedings{milward2003ontology,
  title={{Ontology-based Dialogue Systems}},
  author={Milward, David and Beveridge, Martin},
  booktitle={Proceedings of the 3rd Workshop on Knowledge and reasoning in practical dialogue systems (IJCAI)},
  pages={9--18},
  year={2003},
  organization={Citeseer},
  url={http://citeseerx.ist.psu.edu/viewdoc/summary?doi=10.1.1.136.5607}
}

@article{young2013pomdp,
  title={{{POMDP}-based Statistical Spoken Dialog Systems: {A} Review}},
  author={Young, Steve and Ga{\v{s}}i{\'c}, Milica and Thomson, Blaise and Williams, Jason D},
  journal={Proceedings of the IEEE},
  volume={101},
  number={5},
  pages={1160--1179},
  year={2013},
  publisher={IEEE},
  doi={10.1109/JPROC.2012.2225812}
}

@article{frantzi1999c,
  title={{The {C}-value/{NC}-value Domain-independent Method for Multi-word Term Extraction}},
  author={Frantzi, Katerina T and Ananiadou, Sophia},
  journal={Journal of Natural Language Processing},
  volume={6},
  number={3},
  pages={145--179},
  year={1999},
  publisher={The Association for Natural Language Processing},
  doi={10.5715/jnlp.6.3_145}
}

@inproceedings{nakagawa2002simple,
    title = {{A Simple but Powerful Automatic Term Extraction Method}},
    author = "Nakagawa, Hiroshi  and
      Mori, Tatsunori",
    booktitle = "{COLING}-02: {COMPUTERM} 2002: Second International Workshop on Computational Terminology",
    year = "2002",
    url = "https://aclanthology.org/W02-1407",
}

@inproceedings{wermter2006you,
    title = {{You Can{'}t Beat Frequency (Unless You Use Linguistic Knowledge) {--} A Qualitative Evaluation of Association Measures for Collocation and Term Extraction}},
    author = "Wermter, Joachim  and
      Hahn, Udo",
    booktitle = "Proceedings of the 21st International Conference on Computational Linguistics and 44th Annual Meeting of the Association for Computational Linguistics",
    month = jul,
    year = "2006",
    address = "Sydney, Australia",
    publisher = "Association for Computational Linguistics",
    url = "https://aclanthology.org/P06-1099",
    doi = "10.3115/1220175.1220274",
    pages = "785--792",
}

@inproceedings{yu-etal-2022-unsupervised,
    title = {{Unsupervised Slot Schema Induction for Task-oriented Dialog}},
    author = "Yu, Dian  and
      Wang, Mingqiu  and
      Cao, Yuan  and
      Shafran, Izhak  and
      Shafey, Laurent  and
      Soltau, Hagen",
    editor = "Carpuat, Marine  and
      de Marneffe, Marie-Catherine  and
      Meza Ruiz, Ivan Vladimir",
    booktitle = "Proceedings of the 2022 Conference of the North American Chapter of the Association for Computational Linguistics: Human Language Technologies",
    month = jul,
    year = "2022",
    address = "Seattle, United States",
    publisher = "Association for Computational Linguistics",
    url = "https://aclanthology.org/2022.naacl-main.86/",
    doi = "10.18653/v1/2022.naacl-main.86",
    pages = "1174--1193"
}

@inproceedings{finch-etal-2024-transforming,
    title = {{Transforming Slot Schema Induction with Generative Dialogue State Inference}},
    author = "Finch, James D.  and
      Zhao, Boxin  and
      Choi, Jinho D.",
    editor = "Kawahara, Tatsuya  and
      Demberg, Vera  and
      Ultes, Stefan  and
      Inoue, Koji  and
      Mehri, Shikib  and
      Howcroft, David  and
      Komatani, Kazunori",
    booktitle = "Proceedings of the 25th Annual Meeting of the Special Interest Group on Discourse and Dialogue",
    month = sep,
    year = "2024",
    address = "Kyoto, Japan",
    publisher = "Association for Computational Linguistics",
    url = "https://aclanthology.org/2024.sigdial-1.27/",
    doi = "10.18653/v1/2024.sigdial-1.27",
    pages = "317--324"
}

@article{gruber1995ontologydesign,
title = {{Toward Principles for the Design of Ontologies used for Knowledge Sharing?}},
journal = {International Journal of Human-Computer Studies},
volume = {43},
number = {5},
pages = {907-928},
year = {1995},
issn = {1071-5819},
doi = {https://doi.org/10.1006/ijhc.1995.1081},
url = {https://www.sciencedirect.com/science/article/pii/S1071581985710816},
author = {Thomas R. Gruber}
}

@inproceedings{chamberlin1974sql, author = {Chamberlin, Donald D. and Boyce, Raymond F.}, title = {{SEQUEL: A Structured English Query Language}}, year = {1974}, isbn = {9781450374156}, publisher = {Association for Computing Machinery}, address = {New York, NY, USA}, url = {https://doi.org/10.1145/800296.811515}, doi = {10.1145/800296.811515}, booktitle = {Proceedings of the 1974 ACM SIGFIDET (Now SIGMOD) Workshop on Data Description, Access and Control}, pages = {249–264}, numpages = {16}, location = {Ann Arbor, Michigan}, series = {SIGFIDET '74} }

@inproceedings{huguet-cabot-navigli-2021-rebel-relation,
    title = {{{REBEL}: Relation Extraction By End-to-end Language generation}},
    author = "Huguet Cabot, Pere-Llu{\'i}s  and
      Navigli, Roberto",
    editor = "Moens, Marie-Francine  and
      Huang, Xuanjing  and
      Specia, Lucia  and
      Yih, Scott Wen-tau",
    booktitle = "Findings of the Association for Computational Linguistics: EMNLP 2021",
    month = nov,
    year = "2021",
    address = "Punta Cana, Dominican Republic",
    publisher = "Association for Computational Linguistics",
    url = "https://aclanthology.org/2021.findings-emnlp.204/",
    doi = "10.18653/v1/2021.findings-emnlp.204",
    pages = "2370--2381"
}

@inproceedings{roller-etal-2018-hearst,
    title = {{Hearst Patterns Revisited: Automatic Hypernym Detection from Large Text Corpora}},
    author = "Roller, Stephen  and
      Kiela, Douwe  and
      Nickel, Maximilian",
    editor = "Gurevych, Iryna  and
      Miyao, Yusuke",
    booktitle = "Proceedings of the 56th Annual Meeting of the Association for Computational Linguistics (Volume 2: Short Papers)",
    month = jul,
    year = "2018",
    address = "Melbourne, Australia",
    publisher = "Association for Computational Linguistics",
    url = "https://aclanthology.org/P18-2057/",
    doi = "10.18653/v1/P18-2057",
    pages = "358--363"
}

@inproceedings{macqueen1967some,
  title={{Some Methods for Classification and Analysis of Multivariate Observations}},
  author={MacQueen, James},
  booktitle={Proceedings of the fifth Berkeley symposium on mathematical statistics and probability},
  volume={1},
  pages={281--297},
  year={1967},
  organization={Oakland, CA, USA}
}

@article{ROUSSEEUW198753,
title = {{Silhouettes: A Graphical Aid to the Interpretation and Validation of Cluster Analysis}},
journal = {Journal of Computational and Applied Mathematics},
volume = {20},
pages = {53-65},
year = {1987},
issn = {0377-0427},
doi = {https://doi.org/10.1016/0377-0427(87)90125-7},
url = {https://www.sciencedirect.com/science/article/pii/0377042787901257},
author = {Peter J. Rousseeuw}
}

@inproceedings{budzianowski-etal-2018-multiwoz,
    title = {{{M}ulti{WOZ} - A Large-Scale Multi-Domain {W}izard-of-{O}z Dataset for Task-Oriented Dialogue Modelling}},
    author = "Budzianowski, Pawe{\l}  and
      Wen, Tsung-Hsien  and
      Tseng, Bo-Hsiang  and
      Casanueva, I{\~n}igo  and
      Ultes, Stefan  and
      Ramadan, Osman  and
      Ga{\v{s}}i{\'c}, Milica",
    editor = "Riloff, Ellen  and
      Chiang, David  and
      Hockenmaier, Julia  and
      Tsujii, Jun{'}ichi",
    booktitle = "Proceedings of the 2018 Conference on Empirical Methods in Natural Language Processing",
    month = oct # "-" # nov,
    year = "2018",
    address = "Brussels, Belgium",
    publisher = "Association for Computational Linguistics",
    url = "https://aclanthology.org/D18-1547/",
    doi = "10.18653/v1/D18-1547",
    pages = "5016--5026"
}

@article{heck-etal-2022-robust,
    title = {{Robust Dialogue State Tracking with Weak Supervision and Sparse Data}},
    author = "Heck, Michael  and
      Lubis, Nurul  and
      van Niekerk, Carel  and
      Feng, Shutong  and
      Geishauser, Christian  and
      Lin, Hsien-Chin  and
      Ga{\v{s}}i{\'c}, Milica",
    editor = "Roark, Brian  and
      Nenkova, Ani",
    journal = "Transactions of the Association for Computational Linguistics",
    volume = "10",
    year = "2022",
    address = "Cambridge, MA",
    publisher = "MIT Press",
    url = "https://aclanthology.org/2022.tacl-1.68/",
    doi = "10.1162/tacl_a_00513",
    pages = "1175--1192"
}

@article{raffel2020t5,
author = {Raffel, Colin and Shazeer, Noam and Roberts, Adam and Lee, Katherine and Narang, Sharan and Matena, Michael and Zhou, Yanqi and Li, Wei and Liu, Peter J.},
title = {{Exploring the Limits of Transfer Learning with a unified Text-to-text Transformer}},
year = {2020},
issue_date = {January 2020},
publisher = {JMLR.org},
volume = {21},
number = {1},
issn = {1532-4435},
journal = {J. Mach. Learn. Res.},
month = jan,
articleno = {140},
numpages = {67}
}

@inproceedings{finch-choi-2024-diverse,
    title = {{Diverse and Effective Synthetic Data Generation for Adaptable Zero-Shot Dialogue State Tracking}},
    author = "Finch, James D.  and
      Choi, Jinho D.",
    editor = "Al-Onaizan, Yaser  and
      Bansal, Mohit  and
      Chen, Yun-Nung",
    booktitle = "Findings of the Association for Computational Linguistics: EMNLP 2024",
    month = nov,
    year = "2024",
    address = "Miami, Florida, USA",
    publisher = "Association for Computational Linguistics",
    url = "https://aclanthology.org/2024.findings-emnlp.731/",
    doi = "10.18653/v1/2024.findings-emnlp.731",
    pages = "12527--12544"
}

@book{dennett1987intentional,
  title     = {{The Intentional Stance}},
  author    = {Dennett, Daniel C.},
  year      = {1987},
  publisher = {MIT Press},
  address   = {Cambridge, MA}
}

@inproceedings{razavi2025promptsensitivity,
author = {Razavi, Amirhossein and Soltangheis, Mina and Arabzadeh, Negar and Salamat, Sara and Zihayat, Morteza and Bagheri, Ebrahim},
title = {{Benchmarking Prompt Sensitivity in Large Language Models}},
year = {2025},
isbn = {978-3-031-88713-0},
publisher = {Springer-Verlag},
address = {Berlin, Heidelberg},
url = {https://doi.org/10.1007/978-3-031-88714-7_29},
doi = {10.1007/978-3-031-88714-7_29},
booktitle = {Advances in Information Retrieval: 47th European Conference on Information Retrieval, ECIR 2025, Lucca, Italy, April 6–10, 2025, Proceedings, Part III},
pages = {303–313},
numpages = {11},
location = {Lucca, Italy}
}

@inproceedings{du2021humaneval,
author = {Du, Xueying and Liu, Mingwei and Wang, Kaixin and Wang, Hanlin and Liu, Junwei and Chen, Yixuan and Feng, Jiayi and Sha, Chaofeng and Peng, Xin and Lou, Yiling},
title = {{Evaluating Large Language Models in Class-Level Code Generation}},
year = {2024},
isbn = {9798400702174},
publisher = {Association for Computing Machinery},
address = {New York, NY, USA},
url = {https://doi.org/10.1145/3597503.3639219},
doi = {10.1145/3597503.3639219},
booktitle = {Proceedings of the IEEE/ACM 46th International Conference on Software Engineering},
articleno = {81},
numpages = {13},
location = {Lisbon, Portugal},
series = {ICSE '24}
}

@article{wang2024chain,
  title={{Chain-of-Thought Reasoning without Prompting}},
  author={Wang, Xuezhi and Zhou, Denny},
  journal={arXiv preprint arXiv:2402.10200},
  year={2024}
}

@inproceedings{hudecek-etal-2021-discovering,
    title = {{Discovering Dialogue Slots with Weak Supervision}},
    author = "Hude{\v{c}}ek, Vojt{\v{e}}ch  and
      Du{\v{s}}ek, Ond{\v{r}}ej  and
      Yu, Zhou",
    editor = "Zong, Chengqing  and
      Xia, Fei  and
      Li, Wenjie  and
      Navigli, Roberto",
    booktitle = "Proceedings of the 59th Annual Meeting of the Association for Computational Linguistics and the 11th International Joint Conference on Natural Language Processing (Volume 1: Long Papers)",
    month = aug,
    year = "2021",
    address = "Online",
    publisher = "Association for Computational Linguistics",
    url = "https://aclanthology.org/2021.acl-long.189/",
    doi = "10.18653/v1/2021.acl-long.189",
    pages = "2430--2442"
}

@article{Jacobs1991AdaptiveMixtures,
  author    = {Jacobs, Robert A. and Jordan, Michael I. and Nowlan, Steven J. and Hinton, Geoffrey E.},
  title     = {{Adaptive Mixtures of Local Experts}},
  journal   = {Neural Computation},
  year      = {1991},
  volume    = {3},
  number    = {1},
  pages     = {79--87},
  doi       = {10.1162/neco.1991.3.1.79}
}

@misc{MetaAI2025_Llama4Blog,
  author       = {MetaAI},
  title        = {{The Llama 4 Herd: The Beginning of a New Era of Natively Multimodal AI}},
  howpublished = {Blog post on Meta’s AI site},
  month        = apr,
  year         = {2025},
  url          = {https://ai.meta.com/blog/llama-4-multimodal-intelligence/},
  note         = {Accessed 2025-10-30}
}

@article{qwen3technicalreport,
      title={{Qwen3 Technical Report}}, 
      author={Qwen-Team},
      year={2025},
      eprint={2505.09388},
      archivePrefix={arXiv},
      primaryClass={cs.CL},
      url={https://arxiv.org/abs/2505.09388}, 
      journal={arXiv preprint arXiv:2505.09388},
}

@inproceedings{Shazeer2017OutrageouslyLargeMoE,
  author    = {Shazeer, Noam and Mirhoseini, Azalia and Maziarz, Krzysztof and Davis, Andy and Le, Quoc V. and Hinton, Geoffrey E. and Dean, Jeff},
  title     = {{Outrageously Large Neural Networks: The Sparsely-Gated Mixture-of-Experts Layer}},
  booktitle = {International Conference on Learning Representations (ICLR) 2017},
  year      = {2017},
  url       = {https://openreview.net/forum?id=B1ckMDqlg},
  note      = {Poster presentation}
}

@misc{OpenAI2025_GPT-OSS,
  author       = {OpenAI},
  title        = {{Introducing GPT-OSS}},
  howpublished = {Blog post on OpenAI website},
  month        = {Aug},
  day          = {5},
  year         = {2025},
  url          = {https://openai.com/index/introducing-gpt-oss/},
  note         = {Accessed 2025-10-30}
}

@misc{MetaAI2024_LLaMA31Blog,
  author       = {MetaAI},
  title        = {{Meta Llama 3.1}},
  howpublished = {Blog post on Meta’s AI site},
  month        = jul,
  year         = {2024},
  url          = {https://ai.meta.com/blog/meta-llama-3-1/},
  note         = {Accessed 2025-10-30}
}

@inproceedings{Kim2024ReEx,
  title        = {{Re-Ex: Revising after Explanation Reduces the Factual Errors in LLM Responses}},
  author       = {Kim, Juyeon and Lee, Jeongeun and Chang, Yoonho and Choi, Chanyeol and Kim, Junseong and Sohn, Jy-yong},
  booktitle    = {ICLR 2024 Workshop on Reliable and Responsible Foundation Models},
  year         = {2024},
  url          = {https://openreview.net/pdf?id=tyEWrLVU1b}
}

@article{laban2025llms,
  title={{LLMs get lost in Multi-turn Conversation}},
  author={Laban, Philippe and Hayashi, Hiroaki and Zhou, Yingbo and Neville, Jennifer},
  journal={arXiv preprint arXiv:2505.06120},
  year={2025}
}

@inproceedings{yu-etal-2018-spider,
    title = "{{S}pider: A Large-Scale Human-Labeled Dataset for Complex and Cross-Domain Semantic Parsing and Text-to-{SQL} Task}",
    author = "Yu, Tao  and
      Zhang, Rui  and
      Yang, Kai  and
      Yasunaga, Michihiro  and
      Wang, Dongxu  and
      Li, Zifan  and
      Ma, James  and
      Li, Irene  and
      Yao, Qingning  and
      Roman, Shanelle  and
      Zhang, Zilin  and
      Radev, Dragomir",
    editor = "Riloff, Ellen  and
      Chiang, David  and
      Hockenmaier, Julia  and
      Tsujii, Jun{'}ichi",
    booktitle = "Proceedings of the 2018 Conference on Empirical Methods in Natural Language Processing",
    month = oct # "-" # nov,
    year = "2018",
    address = "Brussels, Belgium",
    publisher = "Association for Computational Linguistics",
    url = "https://aclanthology.org/D18-1425/",
    doi = "10.18653/v1/D18-1425",
    pages = "3911--3921"
}

@inproceedings{stricker-paroubek-2024-shot,
    title = "{A Few-shot Approach to Task-oriented Dialogue Enhanced with Chitchat}",
    author = "Stricker, Armand  and
      Paroubek, Patrick",
    editor = "Kawahara, Tatsuya  and
      Demberg, Vera  and
      Ultes, Stefan  and
      Inoue, Koji  and
      Mehri, Shikib  and
      Howcroft, David  and
      Komatani, Kazunori",
    booktitle = "Proceedings of the 25th Annual Meeting of the Special Interest Group on Discourse and Dialogue",
    month = sep,
    year = "2024",
    address = "Kyoto, Japan",
    publisher = "Association for Computational Linguistics",
    url = "https://aclanthology.org/2024.sigdial-1.50/",
    doi = "10.18653/v1/2024.sigdial-1.50",
    pages = "590--602"
}

@inproceedings{burdisso-etal-2024-dialog2flow,
    title = "{{D}ialog2{F}low: Pre-training Soft-Contrastive Action-Driven Sentence Embeddings for Automatic Dialog Flow Extraction}",
    author = "Burdisso, Sergio  and
      Madikeri, Srikanth  and
      Motlicek, Petr",
    editor = "Al-Onaizan, Yaser  and
      Bansal, Mohit  and
      Chen, Yun-Nung",
    booktitle = "Proceedings of the 2024 Conference on Empirical Methods in Natural Language Processing",
    month = nov,
    year = "2024",
    address = "Miami, Florida, USA",
    publisher = "Association for Computational Linguistics",
    url = "https://aclanthology.org/2024.emnlp-main.310/",
    doi = "10.18653/v1/2024.emnlp-main.310",
    pages = "5421--5440"
}

@inproceedings{agrawal-etal-2024-dialog,
    title = "{Dialog Flow Induction for Constrainable {LLM}-Based Chatbots}",
    author = "Agrawal, Stuti  and
      Pillai, Pranav  and
      Uppuluri, Nishi  and
      Gangi Reddy, Revanth  and
      Li, Sha  and
      Tur, Gokhan  and
      Hakkani-Tur, Dilek  and
      Ji, Heng",
    editor = "Kawahara, Tatsuya  and
      Demberg, Vera  and
      Ultes, Stefan  and
      Inoue, Koji  and
      Mehri, Shikib  and
      Howcroft, David  and
      Komatani, Kazunori",
    booktitle = "Proceedings of the 25th Annual Meeting of the Special Interest Group on Discourse and Dialogue",
    month = sep,
    year = "2024",
    address = "Kyoto, Japan",
    publisher = "Association for Computational Linguistics",
    url = "https://aclanthology.org/2024.sigdial-1.6/",
    doi = "10.18653/v1/2024.sigdial-1.6",
    pages = "66--77"
}

@inproceedings{choubey-etal-2025-turning,
    title = "{Turning Conversations into Workflows: A Framework to Extract and Evaluate Dialog Workflows for Service {AI} Agents}",
    author = "Choubey, Prafulla Kumar  and
      Peng, Xiangyu  and
      Bhagavath, Shilpa  and
      Xiong, Caiming  and
      Pentyala, Shiva Kumar  and
      Wu, Chien-Sheng",
    editor = "Che, Wanxiang  and
      Nabende, Joyce  and
      Shutova, Ekaterina  and
      Pilehvar, Mohammad Taher",
    booktitle = "Findings of the Association for Computational Linguistics: ACL 2025",
    month = jul,
    year = "2025",
    address = "Vienna, Austria",
    publisher = "Association for Computational Linguistics",
    url = "https://aclanthology.org/2025.findings-acl.203/",
    doi = "10.18653/v1/2025.findings-acl.203",
    pages = "3933--3954",
    ISBN = "979-8-89176-256-5"
}

@inproceedings{deng-etal-2024-unveiling,
    title = "{Unveiling the Spectrum of Data Contamination in Language Model: A Survey from Detection to Remediation}",
    author = "Deng, Chunyuan  and
      Zhao, Yilun  and
      Heng, Yuzhao  and
      Li, Yitong  and
      Cao, Jiannan  and
      Tang, Xiangru  and
      Cohan, Arman",
    editor = "Ku, Lun-Wei  and
      Martins, Andre  and
      Srikumar, Vivek",
    booktitle = "Findings of the Association for Computational Linguistics: ACL 2024",
    month = aug,
    year = "2024",
    address = "Bangkok, Thailand",
    publisher = "Association for Computational Linguistics",
    url = "https://aclanthology.org/2024.findings-acl.951/",
    doi = "10.18653/v1/2024.findings-acl.951",
    pages = "16078--16092"
}

@inproceedings{samuel-etal-2025-towards,
    title = "{Towards Data Contamination Detection for Modern Large Language Models: Limitations, Inconsistencies, and Oracle Challenges}",
    author = "Samuel, Vinay  and
      Zhou, Yue  and
      Zou, Henry Peng",
    editor = "Rambow, Owen  and
      Wanner, Leo  and
      Apidianaki, Marianna  and
      Al-Khalifa, Hend  and
      Eugenio, Barbara Di  and
      Schockaert, Steven",
    booktitle = "Proceedings of the 31st International Conference on Computational Linguistics",
    month = jan,
    year = "2025",
    address = "Abu Dhabi, UAE",
    publisher = "Association for Computational Linguistics",
    url = "https://aclanthology.org/2025.coling-main.338/",
    pages = "5058--5070"
}

@article{jiang2023mistral7b,
      title={{Mistral 7B}}, 
      author={Albert Q. Jiang and Alexandre Sablayrolles and Arthur Mensch and Chris Bamford and Devendra Singh Chaplot and Diego de las Casas and Florian Bressand and Gianna Lengyel and Guillaume Lample and Lucile Saulnier and Lélio Renard Lavaud and Marie-Anne Lachaux and Pierre Stock and Teven Le Scao and Thibaut Lavril and Thomas Wang and Timothée Lacroix and William El Sayed},
      year={2023},
      eprint={2310.06825},
      archivePrefix={arXiv},
      primaryClass={cs.CL},
      url={https://arxiv.org/abs/2310.06825}, 
      journal={arXiv preprint: arXiv:2310.06825}
}

\appendix



\onecolumn

\section{Supplementary Information}

\subsection{{\TeQoDO} Prompt} 
\label{appendix:sec:prompt}

\begin{figure}[ht!]
    \scriptsize
    \begin{enumerate}

   \item  You're working with a dialogue system that stores structured data from conversations in an SQLite3 database. The database includes tables covering user intents, system actions, and information about various entities.
    You will be provided with two inputs: the current set of database tables and their names (but not their schemas).
    New dialogue(s) – this contains user and system turns with references to specific intents, actions, or queried information.
    Your task:\\    
    Identify which tables from \{db\_result\_input\} are relevant to the new dialogue(s) based on:
    User intents expressed in the dialogue(s).
    System actions performed in response.
    Specific information or entities being queried or discussed.
    For each relevant table, generate the following SQLite command to inspect its schema:
    PRAGMA table\_info(<table\_name>);
    This will allow you to understand the structure (columns and data types) of the tables you will be working with.
    Do not create or modify tables yet — only inspect existing ones using PRAGMA.
    Make sure to wrap all SQL between ```sql and ``` for it to be parsed.
    
    \item You’ve already examined the database schema using PRAGMA table\_info(...) queries. The schema details are provided, which contain the structure (columns and types) of the current tables in the SQLite3 database.
    Now, based on the same dialogue and the table definitions, your task is to:
    Generate SQL SELECT queries to retrieve:  
    User intents expressed in the dialogue — in general form (e.g., find\_flight, book\_hotel) without including specific parameters (e.g., cities, dates).
    Use the table column schema to determine which table holds this information and write a query to retrieve matching intents: \{db\_result\_input\}\\
    System actions carried out in the dialogue — again in a generalized form (e.g., recommend, confirm, inform).
    Use the appropriate table from above and generate a query to retrieve those action types.
    Information explicitly requested by the user — such as facts about entities (e.g., list of Italian restaurants, hotel prices, flight times), but only if that data is already present in the database.
    Generate SELECT queries from the relevant tables, based on what was asked in the dialogue.
    Do not:
    Create or alter tables (no CREATE, INSERT, or UPDATE);
    Use timestamp fields or session-specific filters;
    Use specific slot values from the dialogue (e.g., exact restaurant names or locations) in the intent or action queries — keep them general.

    \item You’ve already run a set of SELECT queries based on the previous dialogue, and the results of those queries are provided in the following. These results represent all the information currently stored in the database that matches the dialogue. \{db\_result\_input\}
    Your task now is to perform Dialogue State Tracking (DST) by extracting structured information from the dialogue — but only if that information is already present in the database, as confirmed by the query results.\\
    Specifically:
    Use the dialogue to identify user intents, system actions, and information about entities (e.g., preferences, attributes, categories).
    For each matching element, only include it in the tracked state if it appears in the DB results above.
    Represent the extracted state using a table → column → value structure, reflecting exactly how the information maps to the current database.
    Do not:
    Track or infer values that do not exist in the current DB results;
    Use placeholder values or hypothetical interpretations;
    Modify, insert, or extend the database structure in any way.
    Your output should reflect only what is both: Mentioned in the dialogue, and
    Already stored in the database results above.
    
    \item You have already reviewed the current database contents using SELECT queries. The structure and current entries of the database are given in: \{db\_result\_input\}. 
    You’ve also performed Dialogue State Tracking (DST), which revealed the information from the dialogue that is already present in the DB. Now, based on the dialogue and what is missing from the DST results (i.e., what’s not yet in the DB), generate SQL queries (SQLite3 syntax) to bring the database up to date. This will ensure that the dialogue can be successfully handled using only the data in the database.\\
    Your SQL queries should:
    Insert missing user intents into the database using general labels (e.g., book\_train, find\_hotel), as observed in the dialogue and not yet stored according to DB results above.
    Insert missing system actions in generalized form (e.g., inform, offer\_options, confirm\_request) that were present in the dialogue but missing from the DB.
    Insert or update entity information:
    Use INSERT statements if an entity mentioned in the dialogue does not yet exist in the DB.
    Use UPDATE statements if an entity exists but is missing column values (e.g., NULL) that are provided in the dialogue.
    Modify the schema if needed:
    Use ALTER TABLE if the dialogue introduces a new attribute not present in any table based on the current DB results.
    Use CREATE TABLE if a new entity type is mentioned in the dialogue that has no table yet.
    Do not:
    Generate Python code — write only raw SQL queries;
    Use timestamps or session data;
    Update values that are already correctly populated;
    Add user intents or actions with specific slot values from the dialogue — keep them generalized.
    Once the updates are applied, the database should fully support executing and resolving the dialogue based on its contents, so that the user's goal expressed in the dialogue can be successfully fulfilled using only information stored in the database.
    Make sure to have tables for different types of entities, e.g. a restaurants table, etc. and not one table for all entities.

    \end{enumerate}
    \vspace{-0.5cm}
    \caption{Prompts for {\TeQoDO} steps. db\_result\_input is the result of the DB queries from the prior step.}
    \label{fig:appendix:prompt}
\end{figure}
\begin{table*}[h!]
    \centering
    \scriptsize
    \begin{tabular}{ll}
        \textbf{Baseline:} &
        $\mathcal{C}_i = L(d_i), \quad \forall d_i \in \mathcal{D}, \quad \mathcal{O}_i = \mathcal{C}_0 \cup \dots \cup \mathcal{C}_i, \quad \mathcal{O}_0 = \{\varnothing\}$ \\ \hline
        \\
        \textbf{Iterative Baseline:} &
        $\mathcal{C}_i = L(d_i, \mathcal{O}_{i-1}), \quad \mathcal{O}_{i} = \mathcal{O}_{i-1} \cup \mathcal{C}_{i}, \quad \mathcal{O}_0 = \{\varnothing\}$ \\ \hline
        \\
        \textbf{Tracking:} &
        $\mathcal{C}_i = L(d_i, \mathbf{b}_i), \quad \mathbf{b}_i = $ 
        $\mathcal{O}_{i-1} (d_i) = L(d_i, \mathcal{O}_{i-1}, L(d_i, \mathcal{O}_{i-1}))$ 
        , $\quad \mathcal{O}_{i} = \mathcal{O}_{i-1} \cup \mathcal{C}_{i}, \quad \mathcal{O}_0 = \{\varnothing\}$ \\ \hline
        \\
        \textbf{Success:} &
        $\mathcal{C}_i = L(d_i, \mathbf{b}_i)$, \quad  $\mathbf{b}_i = \mathcal{O}_{i-1} (d_i, \text{is\_successful}(d_i)), \quad \mathcal{O}_{i} = \mathcal{O}_{i-1} \cup \mathcal{C}_{i}, \quad \mathcal{O}_0 = \{\varnothing\}$ \\
    \end{tabular}
    \caption{Update formulas for different SQL-based ontology construction prompts. 
    $\mathcal{C}_i$ are the update messages for a dialogue in structured language $L$, where we use SQL. 
    The  dialogues in the dataset are $\mathcal{D} = \{d_0, \dots, d_n\}$. 
    The ontology after dialogue $d_i$ is $\mathcal{O}_i$ and $\mathbf{b}_i$ is the belief state. The final ontology contains $m$ concepts: 
    $\mathcal{O} = \{c_1, \dots, c_m\}$. 
    $\mathbf{b}_i = L(d_i, \mathcal{O}_{i-1})$ is the domain-slot information extracted from the DB for the current dialogue.}
    \label{appendix:tab:TeQoDO_mathtable}
\end{table*}
\newpage
\clearpage

\subsection{Predicted Ontology Excerpts} 
\label{appendix:sec:ontology_excerpt}

\begin{figure}[ht]
    \centering
        \begin{minipage}{0.9\linewidth} 
            \scriptsize 
            \lstset{basicstyle=\ttfamily\scriptsize, breaklines=true}
            \begin{lstlisting}
'restaurants': {'food_type': {'african',
                               'asian',
                               'asian oriental',
                               'brazilian',
                               'british',
                               'chinese',
                               'european',
                               'french',
                               'gallery',
                               'gastropub',
                               'general',
                               'guesthouse',
                               'indian',
                               'international',
                               'italian',
                               'japanese',
                               'korean',
                               'mediterranean',
                               'modern european',
                               'portuguese',
                               'seafood',
                               'spanish',
                               'thai',
                               ...},
                 'name': {'acorn guest house',
                          'anatolia',
                          'ask restaurant',
                          'bangkok city',
                          'bloomsbury restaurant',
                          'cafe jello gallery',
                          'caffe uno',
                          'cambridge chop house',
                          'cambridge lodge',
                          'charlie chan',
                          'chiquito',
                          'city stop',
                          'city stop restaurant',
                          ...},
                 'phone_number': {'01223241387',
                                  '01223308681',
                                  '01223312112',
                                  '01223351707',
                                  '01223354755',
                                  '01223358899',
                                  '01223362433',
                                  '01223365599',
                                  '01223368786',
                                  ...},
                 'price_range': {'cheap',
                                 'expensive',
                                 'moderate',
                                 'moderately priced',
                                 'pricey',
                                 'varied'}}
            \end{lstlisting}
        \end{minipage}
    \caption{JSON representation of \textit{``restaurant''} table predicted via {\TeQoDO} on MultiWOZ test set.}
    \label{fig:ontology_excerpt}
\end{figure}

\begin{figure}[ht]
    \centering
        \begin{minipage}{0.9\linewidth}
            \scriptsize 
            \lstset{basicstyle=\ttfamily\scriptsize, breaklines=true}
            \begin{lstlisting}
'flights': {'airline': {'air france',
                         'alaska airlines',
                         'american airlines',
                         'delta airlines',
                         'generic airline',
                         'n/a',
                         'new airline',
                         'placeholder airline',
                         'southwest airlines',
                         'united airlines'},
             'arrival_city': {'arrival city',
                              'atlanta',
                              'atlanta, ga',
                              'berlin',
                              'chi-town',
                              'chicago',
                              'london',
                              'los angeles',
                              'nairobi',
                              'new york',
                              'new york city',
                              'none',
                              'nyc',
                              'paris',
                              'philadelphia',
                              'philly',
                              'phoenix',
                              'portland',
                              'portland, or',
                              'rio de janeiro',
                              'san diego',
                              'san francisco',
                              'seattle',
                              'sf',
                              'sydney',
                              'toronto',
                              'vancouver',
                              'washington'},
             'arrival_time': {'01:50 pm',
                              ...,
                              'n/a',
                              'none',
                              'tbd'},
             'class': {'n/a', 'economy', 'premium economy'},
             'departure_city': {'atl',
                                ...}
            \end{lstlisting}
        \end{minipage}
    \caption{JSON representation of \textit{``flights''} table predicted via {\TeQoDO} on SGD test set.}
    \label{fig:ontology_excerpt_sgd}
\end{figure}

\newpage
\clearpage

\subsection{Example Dialogues} 
\label{appendix:sec:example_dialogues}

See \Cref{fig:example_dialogues} for the example input dialogues from MultiWOZ test set for the generated queries in \Cref{tab:query_update_qualitative_analysis}.

\begin{figure}[ht]
    \centering
    \begin{minipage}{0.45\textwidth}
        \centering
        \textbf{Dialogue for Similarity Matching Example (multiwoz-test-300)} \\
        \small
        \begin{tabbing}
            \hspace{2cm} \= \hspace{5cm} \= \kill
            \textbf{User}: \> I am looking for a cheap place to stay. \\
            \textbf{System}: \> We have ten cheap options, any preferences. \\
            \textbf{User}: \> We will need free parking. Do any of the options have 4 stars? \\
            \textbf{System}: \> Allenbell matches your requirements, would that work for you? \\
            \textbf{User}: \> I need the hotel type, address and to know if they have wifi. \\
            \textbf{System}: \> Allenbell is a guesthouse, located at 517a Coldham Lane. Yes, they do have wifi. \\
            \textbf{User}: \> I'm also looking for a particular attraction. It's called Trinity College. \\
            \textbf{System}: \> Trinity College's phone number is 01223338400. They are on Trinity Street with postcode CB21TQ. \\
            \textbf{User}: \> Thank you so much for the phone number. Goodbye. \\
            \textbf{System}: \> You're welcome. Have a nice day. \\
        \end{tabbing}
    \end{minipage}
    \hfill
    \begin{minipage}{0.45\textwidth}
        \centering
        \textbf{Dialogue for DST Step Example (multiwoz-test-100)} \\
        \small
        \begin{tabbing}
            \hspace{2cm} \= \hspace{5cm} \= \kill
            \textbf{User}: \> Hi, I'm looking for some places to go in Cambridge. \\
            \textbf{System}: \> Sure. Do you know what you would like to do? \\
            \textbf{User}: \> A park or swimming on the East side. \\
            \textbf{System}: \> There is Cherry Hinton Park and Abbey Pool and Astroturf Pitch. Would you like more information? \\
            \textbf{User}: \> What is the phone number for Abbey Pool and Astroturf Pitch? \\
            \textbf{System}: \> The phone number is 01223902088. Can I help with anything else? \\
            \textbf{User}: \> Could you also tell me Abbey Pool's address? \\
            \textbf{System}: \> The address is Pool Way, Whitehill Road, off Newmarket Road. Can I help you any further? \\
            \textbf{User}: \> I would also like help finding a guesthouse. \\
            \textbf{System}: \> What type of guesthouse are you looking for? \\
            \textbf{User}: \> I'd like to find a cheap 4-star place in the cheap price range, please. \\
            \textbf{System}: \> Which area would you like to stay in? \\
            \textbf{User}: \> The area does not matter. Just whatever you think is best in that criteria. \\
            \textbf{System}: \> In that case, I would recommend the Alexander Bed and Breakfast in the centre. \\
            \textbf{User}: \> Thank you! Do they have free WiFi? \\
            \textbf{System}: \> Yes, they do. Would you like me to book a reservation for you? \\
            \textbf{User}: \> No, thanks. I just need their address and phone number. \\
            \textbf{System}: \> Sure thing. Their phone number is 01223525725. The address is 56 Saint Barnabas Road. \\ Anything else I can do for you? \\
            \textbf{User}: \> No, that will be all. Thank you! \\
            \textbf{System}: \> You're very welcome! Take care! \\
        \end{tabbing}
    \end{minipage}
    \caption{Example dialogues for \Cref{tab:query_update_qualitative_analysis} queries.}
    \label{fig:example_dialogues}
\end{figure}

\newpage
\clearpage

\subsection{General Ontology Prediction Examples}
\label{appendix:subsec:general_examples}

\begin{table}[h!]
\centering
\scriptsize
\begin{tabular}{p{1.5cm}p{8.2cm}p{5.2cm}}
\toprule
\textbf{Dataset} & \textbf{Predicted Example SQL Queries} & \textbf{Ground truth Edge Excerpt} \\
\midrule
arXiv &
\begin{verbatim}
CREATE TABLE IF NOT EXISTS Mathematics (
    id INTEGER PRIMARY KEY,
    category_name TEXT,
    parent_id INTEGER
);

INSERT INTO Mathematics (category_name, parent_id) 
VALUES ('Statistical Analysis Techniques', 1);
\end{verbatim}
&
(Quantitative Finance, Risk Management), (Quantitative Finance, Statistical Finance), (Quantitative Finance, Trading and Market Microstructure), (Statistics, Applications), (Statistics, Computation), (Mathematics, Commutative Algebra), (Mathematics, Algebraic Geometry), (Mathematics, Statistics Theory)
\\
\midrule
Wikipedia &
\begin{verbatim}
CREATE TABLE Theatre (
    id INTEGER PRIMARY KEY,
    category_name TEXT,
    parent_id INTEGER
);

CREATE TABLE Performance (
    id INTEGER PRIMARY KEY,
    category_name TEXT,
    parent_id INTEGER
);

INSERT INTO Theatre (category_name, parent_id) 
VALUES ('Dialect Coaching', 1);
INSERT INTO Theatre (category_name, parent_id) 
VALUES ('Diction Coaching', 1);
INSERT INTO Theatre (category_name, parent_id) 
VALUES ('Dramaturgy', 1);
INSERT INTO Theatre (category_name, parent_id) 
VALUES ('Costume Design', 1);

INSERT INTO Performance (category_name, parent_id) 
VALUES ('Live Performance', 1);
INSERT INTO Performance (category_name, parent_id) 
VALUES ('Theatrical Performance', 1);
\end{verbatim}
&
(Injuries, Wounded and disabled military veterans topics), (Injuries, Healing), (Local government, Seats of local government), (Local government, Unincorporated areas), (Local government, Water management authorities), (Scientific problems, Unsolved problems in astronomy), (Youth health, Sex education), (History of organizations, History of schools), (Design history, Architectural history), (Theatrical occupations, Acting), (Theatrical occupations, Dance), (Theatrical occupations, Dance occupations), (Theatrical occupations, Diction coaches), (Theatrical occupations, Drama teachers)
\\
\bottomrule
\end{tabular}
    \caption{
        Examples comparing predicted SQL queries to ground truth edges, categorised by dataset.
        \label{tab:sql-vs-groundtruth-dataset}
    }
\end{table}

\newpage
\clearpage

\section{Complementary Results}

\subsection{TOD Ontology Results per Nodeclass} 
\label{appendix:sec:per_class_results}


\newcommand*{\firstlinewidth}{2.8cm}

\begin{table*}[h]
    \centering
    \tiny
    \begin{tabular}{m{\firstlinewidth}l|ccc|ccc|ccc}
        \toprule
        Approach & EvalType & \multicolumn{3}{c|}{Literal} & \multicolumn{3}{c|}{Fuzzy} & \multicolumn{3}{c}{Continous} \\
        \midrule
        & & F1 & Precision & Recall & F1 & Precision & Recall & F1 & Precision & Recall \\ 
        \midrule

        \multirow{7}{\firstlinewidth}{Direct Update Baseline} & Micro & 1.15 & 0.94 & 1.49 & 63.57 & 56.75 & 72.24 & 32.29 & 23.97 & 49.46 \\
        & Macro & 2.96 & 3.27 & 16.34 & 37.16 & 31.50 & 88.92 & 19.50 & 19.75 & 62.96 \\
        & Domains & 6.29 & 3.31 & 62.50 & 21.21 & 12.07 & 87.50 & 8.70 & 4.58 & 87.50 \\
        & Slots & 6.17 & 3.73 & 17.86 & 42.74 & 28.09 & 89.29 & 18.40 & 11.11 & 53.57 \\
        & Values & 2.35 & 9.29 & 1.35 & 80.17 & 93.48 & 70.17 & 59.75 & 77.43 & 48.64 \\
        & User intents & 0.00 & 0.00 & 0.00 & 11.87 & 6.32 & 98.59 & 3.16 & 1.62 & 69.01 \\
        & System actions & 0.00 & 0.00 & 0.00 & 29.81 & 17.55 & 99.02 & 7.50 & 4.02 & 56.10 \\

        \midrule

        \multirow{7}{\firstlinewidth}{Iterative Query and Update Baseline} 
        & Micro & 13.76 & 26.03 & 9.35 & 78.03 & 85.58 & 71.71 & 65.84 & 69.44 & 62.60 \\
        & Macro & 15.55 & 18.11 & 16.26 & 69.01 & 65.06 & 79.25 & 47.33 & 42.83 & 61.42 \\
        & Domains & 47.06 & 44.44 & 50.00 & 62.50 & 62.50 & 62.50 & 55.56 & 50.00 & 62.50 \\
        & Slots & 15.38 & 12.00 & 21.43 & 60.61 & 52.63 & 71.43 & 49.35 & 38.78 & 67.86 \\
        & Values & 15.28 & 34.11 & 9.85 & 78.37 & 89.08 & 69.97 & 69.37 & 75.19 & 64.39 \\
        & User intents & 0.00 & 0.00 & 0.00 & 53.97 & 37.57 & 95.77 & 43.45 & 28.10 & 95.77 \\
        & System actions & 0.00 & 0.00 & 0.00 & 89.59 & 83.54 & 96.59 & 18.94 & 22.08 & 16.59 \\

        \midrule

        \multirow{7}{\firstlinewidth}{+ Similarity Matching} & Micro & 17.71 & 34.88 & 11.87 & 75.22 & 89.90 & 64.66 & 70.97 & 79.01 & 64.42 \\
        & Macro & 20.56 & 27.72 & 19.29 & 71.86 & 71.01 & 79.24 & 61.02 & 56.84 & 79.19 \\
        & Domains & 62.50 & 62.50 & 62.50 & 80.00 & 85.71 & 75.00 & 70.59 & 66.67 & 75.00 \\
        & Slots & 19.67 & 18.18 & 21.43 & 63.33 & 59.38 & 67.86 & 59.37 & 52.78 & 67.86 \\
        & Values & 20.62 & 57.89 & 12.54 & 75.32 & 95.17 & 62.32 & 73.81 & 91.03 & 62.06 \\
        & User intents & 0.00 & 0.00 & 0.00 & 55.46 & 39.52 & 92.96 & 30.41 & 18.18 & 92.96 \\
        & System actions & 0.00 & 0.00 & 0.00 & 85.17 & 75.28 & 98.05 & 70.90 & 55.52 & 98.05 \\

        \midrule

        \multirow{7}{\firstlinewidth}{+ Column Value Examples} & Micro & 14.04 & 31.19 & 9.06 & 77.78 & 92.82 & 66.94 & 66.88 & 68.80 & 65.07 \\
        & Macro & 16.79 & 22.21 & 16.19 & 78.67 & 79.10 & 81.42 & 65.03 & 58.04 & 81.02 \\
        & Domains & 50.00 & 50.00 & 50.00 & 93.33 & 100.00 & 87.50 & 82.35 & 77.78 & 87.50 \\
        & Slots & 18.18 & 15.79 & 21.43 & 60.00 & 56.25 & 64.29 & 54.55 & 47.37 & 64.29 \\
        & Values & 15.75 & 45.26 & 9.54 & 77.18 & 95.39 & 64.81 & 67.17 & 72.22 & 62.79 \\
        & User intents & 0.00 & 0.00 & 0.00 & 71.74 & 58.41 & 92.96 & 43.42 & 28.33 & 92.96 \\
        & System actions & 0.00 & 0.00 & 0.00 & 91.12 & 85.47 & 97.56 & 77.67 & 64.52 & 97.56 \\

        \midrule

        \multirow{7}{\firstlinewidth}{+ DST Step} & Micro & 11.16 & 21.80 & 7.50 & 73.57 & 92.61 & 61.02 & 69.47 & 82.16 & 60.18 \\
        & Macro & 9.58 & 13.10 & 8.74 & 72.38 & 74.91 & 74.44 & 60.96 & 59.26 & 73.55 \\
        & Domains & 23.53 & 22.22 & 25.00 & 66.67 & 71.43 & 62.50 & 58.82 & 55.56 & 62.50 \\
        & Slots & 11.76 & 13.04 & 10.71 & 65.38 & 70.83 & 60.71 & 59.26 & 61.54 & 57.14 \\
        & Values & 12.63 & 30.26 & 7.98 & 72.67 & 95.92 & 58.49 & 70.58 & 91.11 & 57.61 \\
        & User intents & 0.00 & 0.00 & 0.00 & 66.67 & 51.97 & 92.96 & 39.64 & 25.19 & 92.96 \\
        & System actions & 0.00 & 0.00 & 0.00 & 90.50 & 84.39 & 97.56 & 76.48 & 62.89 & 97.56 \\

        \midrule
        \multirow{7}{\firstlinewidth}{+ DST and Similarity Matching} & Micro & 7.57 & 15.97 & 4.96 & 74.10 & 83.70 & 66.48 & 68.02 & 73.55 & 63.27 \\
        & Macro & 9.54 & 11.11 & 10.68 & 68.07 & 64.54 & 75.53 & 58.05 & 51.88 & 74.13 \\
        & Domains & 28.57 & 23.08 & 37.50 & 55.56 & 50.00 & 62.50 & 45.45 & 35.71 & 62.50 \\
        & Slots & 10.71 & 10.71 & 10.71 & 58.62 & 56.67 & 60.71 & 54.24 & 51.61 & 57.14 \\
        & Values & 8.40 & 21.75 & 5.21 & 73.64 & 85.93 & 64.42 & 68.52 & 78.20 & 60.97 \\
        & User intents & 0.00 & 0.00 & 0.00 & 64.08 & 48.89 & 92.96 & 44.30 & 29.07 & 92.96 \\
        & System actions & 0.00 & 0.00 & 0.00 & 88.44 & 81.22 & 97.07 & 77.73 & 64.82 & 97.07 \\

        \midrule

        \multirow{7}{\firstlinewidth}{+ DST and Column Value Examples} & Micro & 3.84 & 12.13 & 2.28 & 82.18 & 95.55 & 72.09 & 78.19 & 88.54 & 70.01 \\
        & Macro & 13.15 & 17.84 & 11.52 & 81.86 & 85.80 & 81.46 & 72.15 & 71.00 & 81.01 \\
        & Domains & 42.86 & 50.00 & 37.50 & 85.71 & 100.00 & 75.00 & 80.00 & 85.71 & 75.00 \\
        & Slots & 18.87 & 20.00 & 17.86 & 78.43 & 86.96 & 71.43 & 72.73 & 74.07 & 71.43 \\
        & Values & 4.04 & 19.21 & 2.25 & 81.91 & 98.01 & 70.36 & 79.31 & 94.94 & 68.10 \\
        & User intents & 0.00 & 0.00 & 0.00 & 72.93 & 60.00 & 92.96 & 49.81 & 34.02 & 92.96 \\
        & System actions & 0.00 & 0.00 & 0.00 & 90.29 & 84.03 & 97.56 & 78.90 & 66.23 & 97.56 \\

        \midrule

        \multirow{7}{\firstlinewidth}{+ DST and Column Value Examples and Dialogue Success} & Micro & 9.83 & 32.88 & 5.78 & 76.83 & 92.95 & 65.48 & 73.73 & 87.36 & 63.77 \\
        & Macro & 18.09 & 28.29 & 14.77 & 78.56 & 83.31 & 77.01 & 71.58 & 73.10 & 75.93 \\
        & Domains & 57.14 & 66.67 & 50.00 & 80.00 & 85.71 & 75.00 & 75.00 & 75.00 & 75.00 \\
        & Slots & 22.73 & 31.25 & 17.86 & 66.67 & 80.00 & 57.14 & 62.50 & 75.00 & 53.57 \\
        & Values & 10.56 & 43.53 & 6.01 & 75.78 & 94.26 & 63.36 & 73.30 & 90.61 & 61.54 \\
        & User intents & 0.00 & 0.00 & 0.00 & 75.86 & 64.08 & 92.96 & 57.89 & 42.04 & 92.96 \\
        & System actions & 0.00 & 0.00 & 0.00 & 94.51 & 92.52 & 96.59 & 89.19 & 82.85 & 96.59 \\

        \bottomrule
    \end{tabular}
    \caption{All classes results on MultiWOZ test set.}
    \label{tab:appendix:all_mwoz_test_results_ontology_aggregation}
\end{table*}

\begin{table*}[h]
    \centering
    \tiny
    \begin{tabular}{m{\firstlinewidth}l|ccc|ccc|ccc}
        \toprule
        Approach & EvalType & \multicolumn{3}{c|}{Literal} & \multicolumn{3}{c|}{Fuzzy} & \multicolumn{3}{c}{Continuous} \\
        \midrule
        & & F1 & Precision & Recall & F1 & Precision & Recall & F1 & Precision & Recall \\ 
        \midrule

        \multirow{7}{\firstlinewidth}{Direct Update Baseline} & Micro & 0.96 & 0.67 & 1.69 & 63.05 & 55.08 & 73.73 & 23.12 & 15.50 & 45.47 \\
        & Macro & 2.75 & 2.63 & 17.51 & 41.24 & 35.67 & 90.76 & 19.15 & 18.91 & 71.64 \\
        & Domains & 7.09 & 3.71 & 78.95 & 25.00 & 14.29 & 100.00 & 8.98 & 4.70 & 100.00 \\
        & Slots & 4.22 & 3.01 & 7.10 & 59.95 & 47.06 & 82.58 & 24.91 & 17.39 & 43.87 \\
        & Values & 2.45 & 6.43 & 1.51 & 81.65 & 94.98 & 71.60 & 52.11 & 67.45 & 42.46 \\
        & User intents & 0.00 & 0.00 & 0.00 & 17.73 & 9.73 & 99.60 & 4.77 & 2.45 & 89.33 \\

        \midrule

        \multirow{7}{\firstlinewidth}{Iterative Query and Update Baseline} 
        & Micro & 1.76 & 1.17 & 3.61 & 39.79 & 26.11 & 83.57 & 31.56 & 20.10 & 73.39 \\
        & Macro & 5.67 & 4.10 & 12.30 & 42.60 & 36.69 & 88.99 & 29.48 & 22.36 & 84.23 \\
        & Domains & 20.20 & 12.50 & 52.63 & 61.02 & 45.00 & 94.74 & 35.64 & 21.95 & 94.74 \\
        & Slots & 3.91 & 3.15 & 5.16 & 58.15 & 50.23 & 69.03 & 39.55 & 30.53 & 56.13 \\
        & Values & 4.22 & 4.87 & 3.72 & 82.55 & 82.40 & 82.70 & 61.54 & 53.83 & 71.81 \\
        & User intents & 0.00 & 0.00 & 0.00 & 5.48 & 2.82 & 98.81 & 5.15 & 2.64 & 98.81 \\
        & System actions & 0.00 & 0.00 & 0.00 & 5.83 & 3.00 & 99.66 & 5.53 & 2.84 & 99.66 \\

        \midrule

        \multirow{7}{\firstlinewidth}{+ Similarity Matching} & Micro & 4.43 & 4.55 & 4.31 & 79.02 & 78.97 & 79.08 & 58.46 & 49.34 & 71.70 \\
        & Macro & 5.64 & 4.82 & 8.13 & 73.00 & 64.53 & 87.77 & 48.00 & 37.41 & 84.26 \\
        & Domains & 18.18 & 12.77 & 31.58 & 75.00 & 62.07 & 94.74 & 52.94 & 36.73 & 94.74 \\
        & Slots & 4.68 & 4.86 & 4.52 & 65.82 & 64.60 & 67.10 & 53.61 & 50.28 & 57.42 \\
        & Values & 5.34 & 6.46 & 4.55 & 80.03 & 82.41 & 77.79 & 62.82 & 57.04 & 69.91 \\
        & User intents & 0.00 & 0.00 & 0.00 & 66.05 & 49.51 & 99.21 & 32.26 & 19.26 & 99.21 \\
        & System actions & 0.00 & 0.00 & 0.00 & 78.11 & 64.09 & 100.00 & 38.38 & 23.75 & 100.00 \\

        \midrule

        \multirow{7}{\firstlinewidth}{+ Column Value Examples} & Micro & 2.66 & 3.28 & 2.23 & 81.17 & 82.63 & 79.76 & 61.65 & 56.37 & 68.03 \\
        & Macro & 5.53 & 4.42 & 9.65 & 76.90 & 70.50 & 87.01 & 50.94 & 40.34 & 82.29 \\
        & Domains & 20.78 & 13.79 & 42.11 & 66.67 & 51.43 & 94.74 & 45.57 & 30.00 & 94.74 \\
        & Slots & 3.92 & 3.97 & 3.87 & 65.59 & 65.38 & 65.81 & 50.90 & 47.49 & 54.84 \\
        & Values & 2.97 & 4.31 & 2.26 & 81.17 & 83.82 & 78.69 & 64.07 & 62.20 & 66.06 \\
        & User intents & 0.00 & 0.00 & 0.00 & 84.92 & 75.62 & 96.84 & 45.58 & 29.81 & 96.84 \\
        & System actions & 0.00 & 0.00 & 0.00 & 86.13 & 76.23 & 98.99 & 48.60 & 32.21 & 98.99 \\

        \midrule

        \multirow{7}{\firstlinewidth}{+ DST Step} & Micro & 4.50 & 5.93 & 3.63 & 80.22 & 85.98 & 75.17 & 59.53 & 56.35 & 63.10 \\
        & Macro & 6.69 & 6.05 & 11.38 & 74.47 & 69.61 & 84.67 & 47.55 & 38.88 & 79.12 \\
        & Domains & 21.69 & 14.06 & 47.37 & 72.00 & 58.06 & 94.74 & 41.86 & 26.87 & 94.74 \\
        & Slots & 6.57 & 7.56 & 5.81 & 63.31 & 71.54 & 56.77 & 43.62 & 45.45 & 41.94 \\
        & Values & 5.21 & 8.62 & 3.73 & 80.78 & 89.26 & 73.77 & 62.32 & 63.86 & 60.84 \\
        & User intents & 0.00 & 0.00 & 0.00 & 75.23 & 60.88 & 98.42 & 43.92 & 28.26 & 98.42 \\
        & System actions & 0.00 & 0.00 & 0.00 & 81.04 & 68.28 & 99.66 & 46.05 & 29.94 & 99.66 \\

        \midrule
        \multirow{7}{\firstlinewidth}{+ DST and Similarity Matching} & Micro & 2.70 & 3.42 & 2.23 & 80.05 & 83.84 & 76.59 & 58.63 & 54.87 & 62.94 \\
        & Macro & 3.67 & 3.19 & 7.29 & 73.04 & 66.01 & 86.31 & 44.68 & 36.11 & 79.40 \\
        & Domains & 12.37 & 7.69 & 31.58 & 63.16 & 47.37 & 94.74 & 36.73 & 22.78 & 94.74 \\
        & Slots & 2.76 & 2.96 & 2.58 & 64.92 & 66.00 & 63.87 & 43.17 & 42.50 & 43.87 \\
        & Values & 3.22 & 5.27 & 2.31 & 80.60 & 86.86 & 75.19 & 62.38 & 64.22 & 60.64 \\
        & User intents & 0.00 & 0.00 & 0.00 & 74.55 & 60.00 & 98.42 & 40.06 & 25.15 & 98.42 \\
        & System actions & 0.00 & 0.00 & 0.00 & 81.99 & 69.81 & 99.33 & 41.05 & 25.87 & 99.33 \\

        \midrule

        \multirow{7}{\firstlinewidth}{+ DST and Column Value Examples} & Micro & 3.36 & 4.53 & 2.67 & 81.52 & 85.40 & 77.98 & 62.95 & 59.08 & 67.37 \\
        & Macro & 7.16 & 6.23 & 12.35 & 76.59 & 71.02 & 86.53 & 50.06 & 41.07 & 81.04 \\
        & Domains & 24.69 & 16.13 & 52.63 & 69.23 & 54.55 & 94.74 & 42.86 & 27.69 & 94.74 \\
        & Slots & 7.38 & 8.62 & 6.45 & 67.81 & 72.26 & 63.87 & 49.33 & 51.03 & 47.74 \\
        & Values & 3.75 & 6.39 & 2.66 & 81.78 & 87.55 & 76.73 & 65.96 & 66.51 & 65.43 \\
        & User intents & 0.00 & 0.00 & 0.00 & 78.04 & 65.00 & 97.63 & 44.91 & 29.16 & 97.63 \\
        & System actions & 0.00 & 0.00 & 0.00 & 86.09 & 75.77 & 99.66 & 47.22 & 30.94 & 99.66 \\

        \midrule

        \multirow{7}{\firstlinewidth}{+ DST and Column Value Examples and Dialogue Success} & Micro & 3.44 & 6.58 & 2.33 & 75.45 & 86.81 & 66.71 & 64.83 & 68.25 & 61.73 \\
        & Macro & 10.26 & 10.60 & 12.27 & 79.12 & 79.67 & 81.37 & 59.25 & 53.21 & 78.00 \\
        & Domains & 39.22 & 31.25 & 52.63 & 87.18 & 85.00 & 89.47 & 64.15 & 50.00 & 89.47 \\
        & Slots & 8.40 & 12.05 & 6.45 & 66.42 & 77.59 & 58.06 & 52.94 & 61.54 & 46.45 \\
        & Values & 3.70 & 9.70 & 2.29 & 74.78 & 88.78 & 64.60 & 66.21 & 74.82 & 59.38 \\
        & User intents & 0.00 & 0.00 & 0.00 & 78.39 & 66.21 & 96.05 & 52.77 & 36.38 & 96.05 \\
        & System actions & 0.00 & 0.00 & 0.00 & 88.82 & 80.77 & 98.66 & 60.18 & 43.30 & 98.66 \\

        \bottomrule
    \end{tabular}
    \caption{All classes results on SGD testset.}
    \label{tab:appendix:all_sgd_test_results_ontology_aggregation}
\end{table*}

\newpage
\clearpage





\end{document}